\documentclass[preprint,10pt]{elsarticle}

\usepackage{amssymb}
\usepackage{amsmath}
\usepackage{amsthm}

\usepackage{booktabs}
\usepackage{array}
\usepackage{multirow}
\usepackage[table,dvipsnames]{xcolor}
\usepackage{bm}
\usepackage{url}
\usepackage{xurl}
\usepackage{hyperref}
\usepackage{graphicx}

\graphicspath{{figures/}}
\newtheorem{proposition}{Proposition}
\newtheorem{definition}{Definition}
\newtheorem{corollary}{Corollary}
\definecolor{winteal}{HTML}{0F6E57}
\definecolor{loss}{HTML}{C0453B}
\DeclareMathOperator*{\argmax}{arg\,max}
\newcommand{\AUCwq}{\mathrm{AUC}_{\mathrm{wq}}}
\newcommand{\recipeitem}[1]{%
  \par\medskip
  \begingroup
  \parshape=2
    2em \dimexpr\linewidth-2em\relax
    0pt \linewidth
  \noindent
  \makebox[0pt][r]{\textbullet\hspace{0.6em}}%
  #1\par
  \endgroup
}
\biboptions{sort&compress}

\journal{Information Fusion}

\begin{document}

\begin{frontmatter}

\title{A decodability criterion predicts when hidden-state selection beats majority voting in large language models}

\author[a]{Zhixiang Wang}
\author[a]{Ziliang Hong}
\author[a]{Ulas Bagci\corref{cor}}
\ead{ulas.bagci@northwestern.edu}
\cortext[cor]{Corresponding author.}
\affiliation[a]{organization={Northwestern University}, city={Chicago}, country={United States}}

\begin{abstract}
Combining the answers a large language model (LLM) samples for a question into one decision is a test-time information fusion problem, usually solved by majority voting. Voting is unreliable on difficult questions, where the sampled answers share correlated errors, so the wrong answer can win and drawing more samples makes the decision worse. Selecting a candidate by reading a correctness signal from the model's hidden states is a alternative, but its accuracy varies across models and tasks, and no measure indicates when it can be trusted. Therefore, we propose CASE (Correctness-Axis SElection), a dynamic selection combiner that trains a linear gate on the answer-token hidden state and selects the highest-scoring candidate. Its main contribution is \emph{decodability}, a leakage-free measure of how well the gate ranks a question's correct candidates above its incorrect ones, which predicts whether hidden-state selection will outperform voting. A conventional probe appears accurate only because of question-identity leakage, which vanishes under question-grouped evaluation. On held-out data, decodability predicts the accuracy gain of selection over voting with a Pearson correlation $r=0.75$ and a decision threshold near $\mathrm{AUC}=0.60$. Across general and medical LLMs, CASE improves over voting by up to $19$ points on medium-difficulty questions and $16.8$ points on hard questions. Decodability depends on the aligned knowledge a model must recall, not on its scale, and its prediction transfers to an unseen scientific domain within $3.8$ points. It thus provides a practical criterion, measurable in advance, for choosing between learned selection and majority voting.
\end{abstract}



\end{frontmatter}

\section{Introduction}
Large language models (LLMs) are used where a wrong answer is costly. The standard way to improve reliability at test time is to sample several candidate answers and fuse them into one decision. For LLMs that fusion rule is almost always majority voting over sampled chains of thought (self-consistency \cite{r1}): draw $N$ candidates, return the answer that appears most often. It is training-free, it costs nothing beyond the samples, and it works whenever the correct answer is the plurality of the pool.

That condition fails where reliability matters most. On hard questions the model is wrong more often than right, so the correct answer is a minority of the pool and voting returns the dominant wrong option. The failure is structural, not sampling noise. All candidates come from one model and one knowledge state, so their errors are correlated, and a wrong answer accumulates into a stable consensus. This is the \emph{Byzantine} regime of consensus fusion \cite{r2}, and it worsens with scale: drawing more samples \cite{r8,r9} puts more mass on the modal answer, so more compute buys a more confident error. Escaping it requires a fusion rule that ranks candidates by something other than how often they appear.

Existing alternatives improve the combiner but do not resolve this tension. One line of work reweights candidates using output-space signals, such as confidence-weighted voting, clustered voting, verbalized confidence \cite{r17,r18}, and semantic-entropy or uncertainty-based estimators \cite{r19,r20}. However, these signals are derived from the same generated answers or output distributions and are often imprecise \cite{r15,r16}. Another line of work relies on external judges, including outcome-reward models, process-reward models, trained verifiers, and generative verifiers \cite{r10,r12,r14}. While effective, these approaches require additional supervision and typically incur extra inference cost. A more efficient alternative is to exploit information already encoded in the model. Prior studies suggest that LLM hidden states can linearly encode answer correctness or truthfulness \cite{r3,r4,r5,r6,r7} and may capture task-relevant information not reflected in the output distribution \cite{r22,r23}. Because these activations are already computed during generation, such an internal fusion rule would be almost free, provided it can be made reliable. The question is therefore when that information is reliable enough to guide fusion, given that hidden states are already known to contain some correctness signal.

Yet internal-state selection has shown inconsistent and fragile results, and our first step is to explain why. The natural implementation is straightforward: train a linear probe on a late-layer hidden state, validate it with ordinary cross-validation, and select the candidate with the highest predicted correctness score. This procedure appears strong in validation but does not hold under a stricter evaluation. We show that the apparent success is largely caused by question-identity leakage. Under random-split cross-validation, candidates from the same question can appear in both the training and test folds. The probe can therefore learn to recognize the question, whose candidates are often dominated by one correctness label, rather than learning a transferable correctness signal. When we evaluate the same approach with leakage-free, question-grouped splits, the selection signal collapses to chance-level performance: within-question AUC is 0.502 on LogiQA \cite{r24}, with only a $+0.2$-point gain over a single agent ($p=0.66$). It sharpens the open question: when is correctness internally decodable well enough to yield a useful fusion rule?

It also asks whether a model's internal state is a more reliable indicator of an answer's correctness than the answer itself---and, if so, when. An affirmative answer would move ``knowing when you are right,'' often treated as a property a model simply has or lacks, onto firmer ground: a measurable quantity, present in the activations of some (model, domain) pairs and absent in others, with the difference predictable in advance. It would also place a principled boundary on test-time scaling. Drawing more samples helps precisely when this internal signal exists---additional candidates then convert into reliability---and hurts when it does not, because the samples of a single model share one knowledge state, so without an independent correctness estimate more of them only reinforce the same correlated error. The practical value of ``sample more and aggregate'' therefore hinges on a property one would like to measure before paying for the samples---which, we show, one can.

We answer this question with a broad, leakage-free study of hidden-state selection, which we name CASE (Correctness-Axis SElection), across general and medical LLMs from 1.5B to 70B parameters and benchmarks spanning reasoning, medicine, mathematics, and graduate-level science. The results center on a single predictive relationship: a measurable, leakage-free quantity---the within-question decodability of correctness---forecasts, before deployment, whether hidden-state selection will beat majority voting. On a held-out calibration-to-deployment split it predicts the deployment-set gain at Pearson $r=0.75$ across 13 of the 15 medical model--task settings (two too small to split), above an empirical threshold near $\mathrm{AUC}\approx0.60$; the far tighter in-sample fit ($r=0.96$) we report only as a consistency check, since predictor and outcome are read from the same gate on the same questions. The quantity measures how well a linear gate can rank a question's own candidates by an internal hidden-state score, so the result is a \emph{diagnostic-gated reliability property}: a cheap measurement, taken in advance, that says whether reading correctness from activations is worth doing for a given model and domain. Controlled single-variable experiments further suggest that this signal emerges when the model must recall latent knowledge aligned with the target answer, but is much weaker when the answer can be directly extracted from a provided passage or when the required knowledge is absent. The same decodability measure also predicts out-of-sample behavior in a new non-medical domain. When the signal is present, hidden-state fusion addresses a failure mode that majority voting cannot: as the number of candidates increases, voting can deteriorate when the dominant answer is wrong, whereas hidden-state selection continues to improve, yielding a $+16.8$-point advantage at $N=16$ on hard questions, rising to $+41.3$ on the subset where the correct answer is a minority. Hidden-state selection also matches the performance of a generative verifier, in the setting tested, at negligible added cost, reusing activations already computed during generation. Finally, mechanistic analyses suggest that the signal functions as a shared and transferable readout of answer correctness rather than a causal direction for steering model behavior.

A single pre-deployment measurement thus determines whether hidden-state fusion will outperform voting for a given model and domain. Because the diagnostic is cheap and leakage-free, practitioners can measure it on a small labeled set before deployment. For a given model and domain, this provides an estimate of whether internal fusion is likely to outperform majority voting, and by approximately how much, rather than revealing its failure only after deployment. We make this practical contribution concrete as follows.

\paragraph{Our contributions}
\begin{itemize}
\item The naive late-layer probe fails under leakage-free evaluation; the cause is question-identity leakage in random-split cross-validation. The usable signal lives in the answer-token representation at late layers.
\item A single leakage-free quantity---within-question decodability (AUC)---is an a-priori estimate of the per-instance competence that dynamic selection relies on but has typically estimated only post hoc. It predicts the fusion-over-voting gain ($r=0.96$ in-sample, $0.75$ out of sample) with a derived threshold $\mathrm{AUC}\approx0.60$ (robust to dropping the whole OpenBioLLM family, $r=0.90$), and it governs not only our selector but any hidden-state score---the same threshold predicts three output-space selectors. One stylized model derives this law, its anti-Condorcet $N$-scaling, and the leakage (Section~\ref{sec:analysis}).
\item Four single-variable controls point to aligned latent knowledge as the likely driver---over domain, scale, or architecture. A same-architecture general model and specialist fall on opposite sides of the threshold, and at a fixed late layer removing the supporting passage (closed- vs.\ open-book) restores the recall-based decodability that is otherwise absent (Section~\ref{sec:controls}).
\item Where the signal is present, CASE improves with the candidate count while voting collapses ($+16.8$ pp on hard questions at $N=16$). It matches a generative verifier at negligible cost.
\item The correctness signal is a shared, transferable readout direction (cosine up to 0.93), and we report two dissociations: it is decodable yet does not causally steer (a readout, not a lever), and on closed-book recall the internal readout ranks correctness while output confidence anti-correlates with it.
\end{itemize}

The rest of the paper is organized as follows. Section~\ref{sec:related} situates the problem within decision fusion, test-time compute aggregation, verifiers, and internal-state probing. Section~\ref{sec:method} formalizes hidden-state selection and the leakage-free decodability diagnostic; Section~\ref{sec:analysis} analyzes why decodability governs the fusion gain, deriving the law, its $N$-scaling, and the leakage from one model; Section~\ref{sec:setup} describes the models and benchmarks. Section~\ref{sec:results} presents the main results: the leakage diagnosis (\ref{sec:leakage}), when internal fusion beats voting (\ref{sec:when}--\ref{sec:controls}), the decodability law (\ref{sec:law}), \emph{Byzantine} $N$-scaling (\ref{sec:nscale}), the mechanism (\ref{sec:mech}), cost and cross-domain generality (\ref{sec:cost}), and a comparison against near-free output-space selectors (\ref{sec:cheap}). Section~\ref{sec:recipe} translates the law into a deployment recipe, and Sections~\ref{sec:discussion}--\ref{sec:conclusion} provide the discussion and conclusion.

\section{Related work}\label{sec:related}

\subsection{Classifier and decision fusion}
Combining multiple weak decisions into a stronger one is a central problem in information fusion, including majority and weighted voting, stacked generalization \cite{r39}, mixtures of experts \cite{r40}, and a broad family of confidence-based and trainable combiners \cite{r37}. 
Two classical themes frame our problem. First, simple voting is effective mainly when errors are sufficiently independent and unbiased; when correlated bias pushes many base decisions toward the same wrong answer, voting can fail. This is the regime formalized in the \emph{Byzantine}-fault literature \cite{r2} and also described in ensemble theory as a consequence of low base-learner diversity \cite{r37}. Second, an effective combiner requires a reliable estimate of candidate quality, and the key design question is where this estimate should come from: held-out validation data, external supervision, or the base learners' own confidence scores. For LLM answer fusion, the base decisions are samples from a single model, so their errors are correlated through the model's knowledge state, and the natural quality estimate is the model's own, often miscalibrated, confidence. We study an internal alternative: a correctness estimate read directly from hidden activations, and ask when it is reliable enough to serve as a fusion weight. In the classifier-fusion taxonomy, this places CASE as a trainable, measurement-level combiner of the \emph{dynamic-selection} type (dynamic classifier/ensemble selection, DCS/DES \cite{r53,r54}): rather than combining all candidates, it selects the single one its internal competence estimate ranks highest, approximating the oracle selector that would pick a correct candidate whenever the pool contains one. Where classical DCS/DES estimates this local competence post hoc from a validation neighborhood \cite{r53,r54}, CASE reads it directly from the model's own residual stream and, distinctively, can certify its quality \emph{before} deployment through the decodability diagnostic. The within-question decodability is then precisely an estimate of that per-item competence, and the law states when the estimate is good enough for selection to beat consensus---a leakage-free, a-priori-measurable instance of the classical condition under which a trainable combiner beats a fixed consensus rule.

\subsection{Decision fusion of stochastic LLM outputs}
Self-consistency \cite{r1} fuses multiple chains of thought by majority vote and is the standard test-time fusion baseline for LLMs. Repeated sampling and test-time compute scaling \cite{r8,r9} motivate drawing larger candidate pools, which can improve accuracy on easy inputs but also worsen the \emph{Byzantine} fault for voting on hard inputs; input-adaptive methods instead learn how much computation each input warrants \cite{r56}, a budgeting question orthogonal to the fusion rule we study; voting accuracy can even be non-monotone in the number of calls \cite{r46}. We show a stronger effect: in the \emph{Byzantine} regime voting collapses while an internal selector keeps improving (Section~\ref{sec:nscale}). Several refinements reweight or filter candidates, including weighted self-consistency, answer clustering, Borda-count ranked voting over self-certainty scores \cite{r42}, and confidence-weighted aggregation, but their weights are still derived from the output distribution. Our work instead asks when an internal signal, not tied to output frequency, provides a better fusion rule than counting answers, and quantifies when it does.

\subsection{Verifiers and reward models}
A second route to better candidate selection is to train an external judge. Outcome- and process-reward models, as well as trained or generative verifiers \cite{r10,r12,r14}, rank or reweight candidates and can be effective, but they require labeled outcome or process data and add a full forward pass, often several, per candidate at inference. Generative verifiers \cite{r14} in particular recast verification as next-token prediction and provide a competitive, training-light baseline. We treat a training-free generative self-verifier as our strong baseline and compare both accuracy and inference cost with the near-free hidden-state probe, positioning internal fusion as a low-cost complement to, rather than a replacement for, trained verifiers (Section~\ref{sec:cost}).

\subsection{Confidence, calibration and uncertainty}
Output-level confidence is a natural fusion weight but is systematically miscalibrated \cite{r15,r16}; verbalized confidence \cite{r17,r18} and semantic-entropy / uncertainty methods \cite{r19,r20} extract better signals from the output distribution and detect hallucinations \cite{r20}. The closest competitor to internal selection is self-certainty \cite{r42}, a near-free best-of-$N$ selector read from the output distribution that, like our probe, costs almost nothing and improves with the candidate count. The two differ only in what they read---the output distribution versus an internal estimate from the residual stream---and we compare against it and related output-space selectors directly, not only against voting and an expensive verifier (Section~\ref{sec:cheap}). Our decodability diagnostic then provides a per-(model, domain) test for when the internal estimate is trustworthy enough to fuse on.

\subsection{Internal-state probing and the geometry of truth}
Linear probes can recover substantial information from intermediate activations \cite{r22}, and prior work shows that LLMs encode truth or falsehood directions that can be found with \cite{r3} or without \cite{r4} supervision, often in an approximately linear form \cite{r5,r23}. These signals tend to concentrate at answer tokens and late layers \cite{r6}, and related directions have been used for inference-time interventions to elicit more truthful generations \cite{r7}. We connect this readout view to answer fusion. Specifically, we quantify when a correctness direction can serve as a fusion weight, show that it is shared and transferable across benchmarks, and demonstrate that, unlike truthfulness directions used for intervention \cite{r7}, the correctness direction studied here acts as a readout rather than a causal lever. Closest to our method are recent hidden-state readouts for candidate selection: lightweight latent verifiers that score correctness from activations and plug into best-of-$N$ and self-consistency (LiLaVe \cite{r49}), token-level hidden-state reward models for best-of-$N$ (SWIFT \cite{r50}), latent process-level scorers for reasoning trajectories (TrajSelector \cite{r51}), and latent-trajectory temporal signals that guide answer selection across sampled generations \cite{r57}; hidden-state probes have likewise been used to select correct answers from sampled candidates \cite{r44} and to verify a reasoning model's own intermediate answers \cite{r45}. We contribute the quantity that governs when \emph{any} such selector helps: a leakage-free decodability score that predicts the downstream fusion-over-voting gain, with a threshold and an out-of-sample test. Two works bear directly on our claims. Cho et al. \cite{r48} map the geometric structure of correctness representations and report both a linear-versus-nonlinear probe comparison and an internal-versus-output-method comparison that parallel our Supplementary Table~S9 and Section~\ref{sec:cheap}; they also report a causal-steering result more positive than ours (Section~\ref{sec:mech}), a discrepancy we attribute to differing intervention protocols (steering norm, layer set, and difficulty targeting) and leave to future work. Seo et al. \cite{r47} show that apparent correctness-prediction gains can be inflated by question-side shortcuts---the effect our leakage audit (Section~\ref{sec:leakage}) isolates. Finally, semantic-entropy probes \cite{r52} are the direct precedent for our efficiency framing---a cheap hidden-state readout that replaces an expensive uncertainty computation---here repurposed from hallucination detection to fusion weighting. Our contribution remains distinct: we link a leakage-free decodability score to the fusion-over-voting gain, establish its threshold, and turn it into an a-priori deployment criterion.

\section{Method}\label{sec:method}

\subsection{Problem formulation}
\begin{figure}[t]\centering
\includegraphics[width=\linewidth,
trim={0bp 5bp 0bp 0bp},
clip]{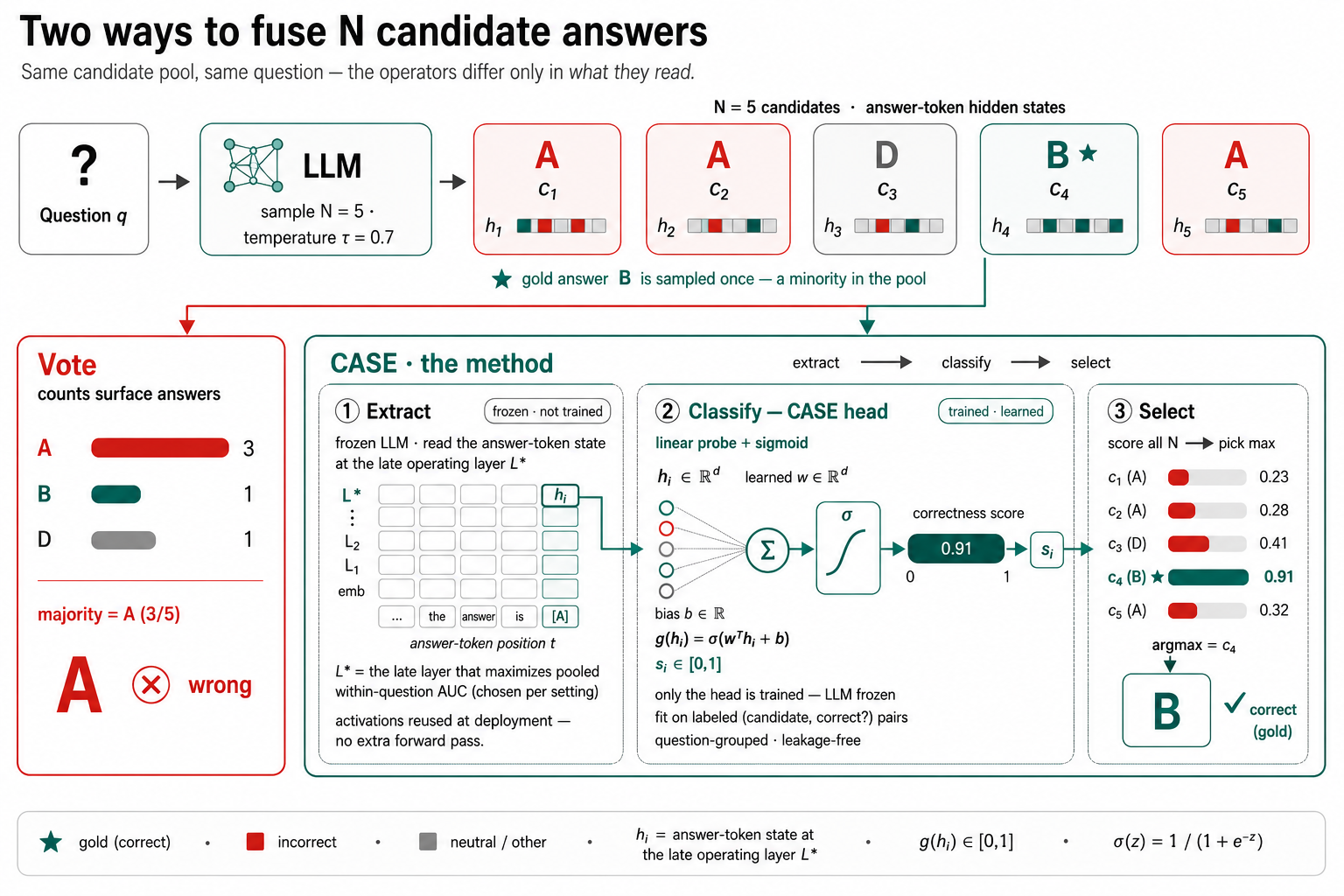}
\caption{Two ways to fuse $N$ candidate answers. Majority voting counts surface answers and fails when the correct answer is a minority; hidden-state selection (CASE) reads an internal correctness score from each candidate's answer-token representation and returns the highest-scoring candidate.}
\label{fig:method}
\end{figure}

Given a question $q$ and $N$ candidates $\{c_1,\dots,c_N\}$ sampled from an LLM at temperature $\tau$, an answer-fusion operator returns a single answer. Writing $\mathrm{ans}(c)$ for the surface answer of candidate $c$, the three operators we compare---the single-agent baseline, majority voting, and hidden-state selection (CASE)---are
\begin{equation}
\begin{aligned}
\hat a_{\text{single}}
&= \mathrm{ans}(c_1),
\qquad
\hat a_{\text{vote}}
= \argmax_{a}
\sum_{i=1}^{N}
\mathbb{1}\!\left[\mathrm{ans}(c_i)=a\right],
\\
\hat a_{\text{CASE}}
&= \mathrm{ans}\!\left(
\argmax_{i} g(h_i)
\right).
\end{aligned}
\end{equation}
where the gate $g$ maps the internal representation $h_i$ of candidate $c_i$ to a correctness score. Voting and CASE use the same candidate pool and differ only in how they fuse it---by surface-answer frequency versus by an internal correctness readout (Fig.~\ref{fig:method})---which isolates the value of the internal signal from the value of sampling itself.

\subsection{Hidden-state extraction}
We extract the signal from the answer-token position---the position that produces the final answer (the option letter for multiple-choice tasks, the final answer token for free-response). For each completed candidate we run one teacher-forced forward pass and record, at every layer, the residual-stream activation at that position---the hidden state from which the answer token is predicted. This calibration pass records all layers at once; at deployment the operating-layer activation is already produced during the generation of each candidate, so scoring adds no forward pass. We compared the answer-token position against two natural alternatives---the last-token activation and the mean over the final $K$ tokens (`mean-$K$')---and it carries the correctness signal substantially better than either; mean-$K$ in particular averages the signal away, leaving every layer at chance on LogiQA. This matches reports that truthfulness concentrates at answer tokens and late layers \cite{r3,r6,r7}. We therefore operate at the late layer that maximizes pooled within-question decodability (Section~\ref{sec:diag}; Section~\ref{sec:law} verifies that selecting this layer on a strictly held-out split does not change our findings); the operating layer for each of the 15 settings is listed in Supplementary Table~S5.

\subsection{Selection gate}
The gate $g$ is a simple linear model: we standardize activations per feature and fit an $\ell_2$-regularized logistic regression (scikit-learn, $C=1.0$) on $(h,\text{correctness})$ pairs,
\begin{equation}
g(h)=\sigma(w^\top h + b),\qquad \sigma(z)=\frac{1}{1+e^{-z}}.
\end{equation}
Two design choices matter. First, the gate is deliberately linear. A sweep over discriminators (logistic, MLP-256, RBF-SVM) shows that the limiting factor is the representation---the layer, token position, and model---not classifier capacity; larger discriminators merely overfit the small per-setting data (best Qwen-7B cell: logistic 0.596 $\geq$ MLP 0.537 $\geq$ SVM 0.535; Supplementary Table~S9). Second, all training and evaluation use 5-fold GroupKFold with the question as the grouping key, so no question appears in both training and test folds; per-feature standardization and the gate are fit on the training folds only. This removes the leakage that inflates the naive random-split implementation. CASE then returns the candidate with the highest out-of-fold gate score.

\subsection{The decodability diagnostic}\label{sec:diag}
Our central quantity is the within-question ranking AUC. For each question with both correct and incorrect candidates, we compute the ROC-AUC of the gate score against candidate correctness---equivalently the Mann--Whitney $U$ statistic, with tied scores counted as $\tfrac12$---then average over questions:
\begin{align}
\mathrm{AUC}_q &= \Pr\!\big[\,g(h_i)>g(h_j)\ \big|\ y_i=1,\ y_j=0\,\big],\quad i,j\in C_q,\\
\AUCwq &= \frac{1}{|Q^\ast|}\sum_{q\in Q^\ast}\mathrm{AUC}_q,\qquad Q^\ast=\{q: 0<\bar y_q<1\},
\end{align}
where $y_i\in\{0,1\}$ is candidate correctness and $\bar y_q$ its per-question mean. This metric has the two properties we need. It is leakage-free by construction: every candidate being ranked belongs to the same question, so question identity cannot help. It also measures exactly what selection requires: whether correctness can be read out well enough to rank one question's own candidates. This differs from grouped cross-validation accuracy, which still aggregates performance across questions. We report this quantity in two forms that differ only in which questions enter $Q^\ast$. The \emph{pooled} within-question AUC averages over all mixed-correctness questions and is what we maximize when selecting the operating layer; the \emph{medium-bin} within-question AUC restricts $Q^\ast$ to medium-difficulty questions (the difficulty bins are defined below) and is the axis on which the law of Section~\ref{sec:law} is fit. Unless qualified, ``within-question AUC'' denotes the pooled form; the law, its slope, and its threshold are stated on the medium-bin form. Section~\ref{sec:law} shows that this metric predicts the downstream fusion gain very strongly, supporting its use as a deployment diagnostic.

\subsection{Baselines and metrics}
We compare CASE against three baselines: (i) the single agent (return $c_1$); (ii) majority voting, with ties broken uniformly at random; and (iii) a zero-training generative self-verifier that prompts the same model ``Is answer X correct? Yes/No'' and ranks candidates by $P(\text{Yes})$ (the P(True) selector \cite{r15}). Because internal selection is valuable only if near-free, we also compare against equally cheap output-distribution selectors---self-certainty \cite{r42}, sequence log-probability, and predictive entropy---read from the same forward pass (Section~\ref{sec:cheap}); semantic-entropy selection \cite{r19} reduces to majority voting on multiple-choice and is not a separate baseline. We further compare against two recent hidden-state selector designs, reimplemented on our pools: a LiLaVe-style shallow-tree latent verifier \cite{r49} and a SWIFT-style token-pooled linear readout \cite{r50} (Section~\ref{sec:cheap}). We bin difficulty by the per-question accuracy over the candidate pool,
\begin{equation}
A_q=\frac1N\sum_{i=1}^{N}\mathbb{1}[c_i\text{ correct}],\qquad
\operatorname{bin}(q)=\begin{cases}\text{easy},& A_q\ge0.6,\\ \text{medium},& 0.35\le A_q<0.6,\\ \text{hard},& A_q<0.35;\end{cases}
\end{equation}
to avoid circularity, we verify that estimating difficulty on a held-out split of candidates leaves the bins essentially unchanged. Our main metric is the medium-difficulty selection gain
\begin{equation}
\Delta=\operatorname{acc}(\hat a_{\text{CASE}})-\operatorname{acc}(\hat a_{\text{vote}})\quad\text{on medium-difficulty questions,}
\end{equation}
the regime with room to improve, averaged over 200 random $N=4$ draws per question, with 95\% bootstrap confidence intervals (3000 resamples over questions) and two-sided $p$-values. Because each per-question gain is itself averaged over 200 draws, we checked that resampling those draws matters little: a two-level bootstrap over both questions and draws (four representative settings) changes the interval widths by at most $0.5$ pp and flips no significance verdict, leaving question sampling dominant. The $N$-scaling analysis instead draws real candidates without replacement from the pool, on the minority-correct subset, so its curves reflect genuine candidate growth rather than resampling. Because we report many per-setting tests, we treat individual $p$-values as descriptive and rest the conclusions on the consistency of the effect across independent settings and on the across-model law (Section~\ref{sec:law}).

\section{Analysis: why decodability governs the fusion gain}\label{sec:analysis}
A single stylized model derives the three central phenomena---the decodability law and its threshold, the \emph{Byzantine} $N$-scaling, and the cross-validation leakage that defeats the naive probe. We state the model once (Definition~\ref{def:model}) and derive each in turn; full proofs are in Supplementary Section~S11.

\begin{definition}[stylized correctness-score model]\label{def:model}
For a question whose candidate pool has correct-fraction $A$ (its per-question accuracy, Section~\ref{sec:diag}), we model the gate score of a sampled correct candidate as an independent draw from a distribution $F_1$ and of an incorrect candidate from $F_0$---the randomness is over the sampled candidates, not the deterministic gate---where $(F_0,F_1)$ form a location family with separation $\delta\in\mathbb{R}$. The within-question decodability is the probability that a correct candidate outscores an incorrect one,
\begin{equation}
a \;=\; \Pr[\,S_1>S_0\,],\qquad S_1\sim F_1,\ \ S_0\sim F_0,
\end{equation}
which increases with $\delta$: from $a<\tfrac12$ when $\delta<0$, through $\tfrac12$ at $\delta=0$ (indistinguishable), to $1$ as $\delta\to\infty$ (perfect separation). Candidates may still concentrate on one wrong option---the correlated-error premise that defeats voting---while their gate scores remain conditionally independent given correctness. CASE returns the argmax-score candidate; majority voting returns the plurality answer, correct with a gate-independent probability $V$.
\end{definition}

Writing $M_1$ and $M_0$ for the largest gate score among the correct and the incorrect candidates, CASE is correct exactly when $M_1>M_0$:
\begin{equation}
\Pr(\text{CASE correct})=\Pr[\,M_1>M_0\,],\qquad M_1=\max_{i\in\text{corr}}g(h_i),\quad M_0=\max_{j\in\text{inc}}g(h_j).
\end{equation}

\paragraph{The law and its threshold}
\begin{proposition}[monotonicity]
Under Definition~\ref{def:model}, $\Pr(\text{CASE correct})$ increases monotonically in the decodability $a$, from the pool accuracy $A$ at $a=\tfrac12$ (an uninformative gate ranks the pool at random) to $1$ as $a\to1$ (perfect separation surfaces a correct candidate whenever the pool contains one, which holds with probability approaching $1$ as $N$ grows):
\begin{equation}
\Pr(\text{CASE correct};\,a=\tfrac12)=A,\qquad \Pr(\text{CASE correct};\,a\to1)\to1.
\end{equation}
\end{proposition}

\begin{proposition}[decodability threshold]
The selection gain over voting,
\begin{equation}
G(a)=\Pr(\text{CASE correct};a)-V,\qquad G(\tfrac12)=A-V,
\end{equation}
is increasing in $a$ and crosses zero at a threshold $a^\ast$ whose position is set by the sign of $V-A$: $a^\ast>\tfrac12$ when voting beats a random pick of the pool ($V>A$), and $a^\ast<\tfrac12$ when it does not ($V<A$).
\end{proposition}

Propositions~1--2 explain the law directly. (i)~The gain rises with decodability---the positive, tight relationship of Section~\ref{sec:law}. (ii)~At medium difficulty the sampled wrong answers are spread across options, so voting is only marginally better than a random pick ($V\gtrsim A$) and the threshold sits just above $\tfrac12$---matching the empirical $\mathrm{AUC}\approx0.60$. The precise value depends on the difficulty distribution and is measured, not derived (Fig.~\ref{fig:law}b).

The threshold applies to medium difficulty; the hard regime requires separate treatment, which the same model provides. Proposition~2 places the crossing at $a^\ast<\tfrac12$ whenever $V<A$---precisely the \emph{Byzantine} regime, where sampled errors concentrate on one wrong option and voting does worse than a random pick of the pool. The model therefore predicts that on hard, \emph{Byzantine} inputs even weakly decodable (sub-threshold) settings should gain, which is what we observe systematically: across the 15 medical settings the hard-bin CASE$-$vote gain is positive in every well-powered case ($14/14$ with a hard bin of adequate size; the five settings with archived candidate pools span $+5.1$ to $+13.4$ pp, Table~S1; full per-setting values in the released per-run logs), essentially independent of the medium-bin AUC that governs the law (the two below-threshold graduate-chemistry settings likewise gain $+6.2$ and $+7.7$ pp; Section~\ref{sec:cost}). The $\mathrm{AUC}\approx0.60$ threshold thus governs medium difficulty; the hard/\emph{Byzantine} regime has no useful threshold. There, voting has already collapsed, so escaping the plurality (even a single candidate does) already helps almost regardless of decodability, and the internal signal adds beyond this only where $a>\tfrac12$---a split we carry into the deployment recipe (Section~\ref{sec:recipe}).

\paragraph{The anti-Condorcet collapse and its remedy ($N$-scaling)} The same model predicts what happens as the candidate count grows.
\begin{proposition}[$N$-scaling]
In the \emph{Byzantine} regime---the correct answer a minority ($A<\tfrac12$) with a modal wrong option, but decodable ($a>\tfrac12$)---and for light-tailed gate scores (the extreme-value regularity of Supplementary Section~S11, satisfied e.g.\ by Gaussian logits), as the candidate count $N$ grows, $\Pr(\text{CASE correct})\to1$ while $\Pr(\text{vote correct})\to0$; the two provably diverge.
\end{proposition}
The intuition is a contest of extremes: more candidates give CASE more correct candidates from which to take a maximal score, and because correct scores are shifted upward ($a>\tfrac12$) that maximum increasingly wins; voting, by contrast, concentrates ever more mass on the modal---wrong---option. This is the Condorcet jury theorem and its negative branch: when the per-candidate probability of the correct option exceeds that of every wrong option, majority voting converges to certainty as $N$ grows; but in the \emph{Byzantine} regime, where a wrong option is modal, the same theorem drives voting toward \emph{zero} (the anti-Condorcet branch), and only a competence-weighted rule---here the internal correctness readout---can still improve with $N$. This is the divergence measured in Section~\ref{sec:nscale} (CASE $20.4\to47.3\%$, voting $20.4\to6.0\%$ over $N=1\to16$; Fig.~\ref{fig:nscale}a); finite $N$ gives the partial recovery observed there rather than the asymptotic limit. Below the threshold ($a<\tfrac12$) the correct maximum no longer wins and no recovery occurs.

\paragraph{Leakage of the naive probe}
\begin{proposition}[leakage]
Under random-split cross-validation, a classifier that recognizes the question and predicts its majority label attains accuracy at least $\mathbb{E}_q[\max(\pi_q,\,1-\pi_q)]$ using question identity alone, where $\pi_q$ is the question's correct-fraction. This exceeds both chance and the within-question ranking AUC whenever the questions are label-imbalanced; grouping the folds by question withholds the test question and removes the term.
\end{proposition}
This question-identity shortcut is the inflated $0.610$ of Section~\ref{sec:leakage} (the split-vs-grouped comparison is Supplementary Table~S6); question-grouped evaluation exposes the genuine within-question decodability ($0.502$, chance). The leakage gap is thus not noise but a predictable artifact of the split, and $\AUCwq$---which conditions on the question by construction---is the leakage-free quantity Propositions~1--3 require.

\paragraph{Connection to fusion theory}
\begin{corollary}[learned combiner vs.\ consensus]
By Proposition~2, the internal combiner CASE outperforms majority voting on a given (model, domain) exactly when its within-question decodability exceeds the threshold $a^\ast$ set by $V-A$; below it, consensus fusion is preferable. The decodability AUC therefore plays a role analogous to the classical accuracy--diversity condition, giving a per-(model, domain) test for when a trainable combiner beats consensus fusion (Section~\ref{sec:discussion}).
\end{corollary}

\section{Experimental setup}\label{sec:setup}
\textbf{Models.} General-purpose: Qwen2.5-\{1.5, 3, 7, 14\}B-Instruct and Qwen2.5-Math-7B \cite{r29}, and Llama-3-8B-Instruct \cite{r30}. Medical specialists: OpenBioLLM-8B/70B \cite{r31}, BioMistral-7B \cite{r32}, Med42-8B \cite{r33}, MMed-Llama-3-8B \cite{r35}, meditron-7B \cite{r34}. This model panel covers three axes---scale (1.5--70B), specialization (general vs.\ medical), and alignment strength within the medical group---and includes a general and a specialist model sharing the Llama-3-8B architecture, which supports the controls in Section~\ref{sec:controls}. All models run in fp16; the 70B model is sharded across GPUs via device-map.

\textbf{Benchmarks.} LogiQA (logical reasoning) \cite{r24}; MedQA (USMLE four-option) \cite{r25}; MedMCQA \cite{r26}; PubMedQA \cite{r27} in both open-book (the supporting abstract is provided) and closed-book (the abstract is removed) forms; and MATH-500 (competition mathematics) \cite{r28} and GSM8K (grade-school mathematics) \cite{r10}; and GPQA (graduate-level physics and chemistry) \cite{r43} as an out-of-domain, non-medical knowledge benchmark (Section~\ref{sec:cost}). The open-/closed-book pair on identical PubMedQA questions is the single-variable control that isolates latent-knowledge recall. For each question we sample 16--20 candidates by nucleus sampling ($\tau=0.7$, top-$p=0.95$, up to 512 new tokens) and extract hidden states as in Section~\ref{sec:method}.

\textbf{Implementation.} Candidate generation, extraction, and analysis run on NVIDIA A6000 and A100 GPUs; gates and statistics use scikit-learn with bootstrap resampling. All probe quality and selection metrics are leakage-free (question-grouped). Code, configurations, and per-run logs are available for review (see Code availability).

\section{Results}\label{sec:results}

\subsection{The naive implementation is a leakage artifact}\label{sec:leakage}
\begin{figure}[t]
\centering
\includegraphics[width=\linewidth,
trim={0bp 10bp 0bp 0bp},
clip]{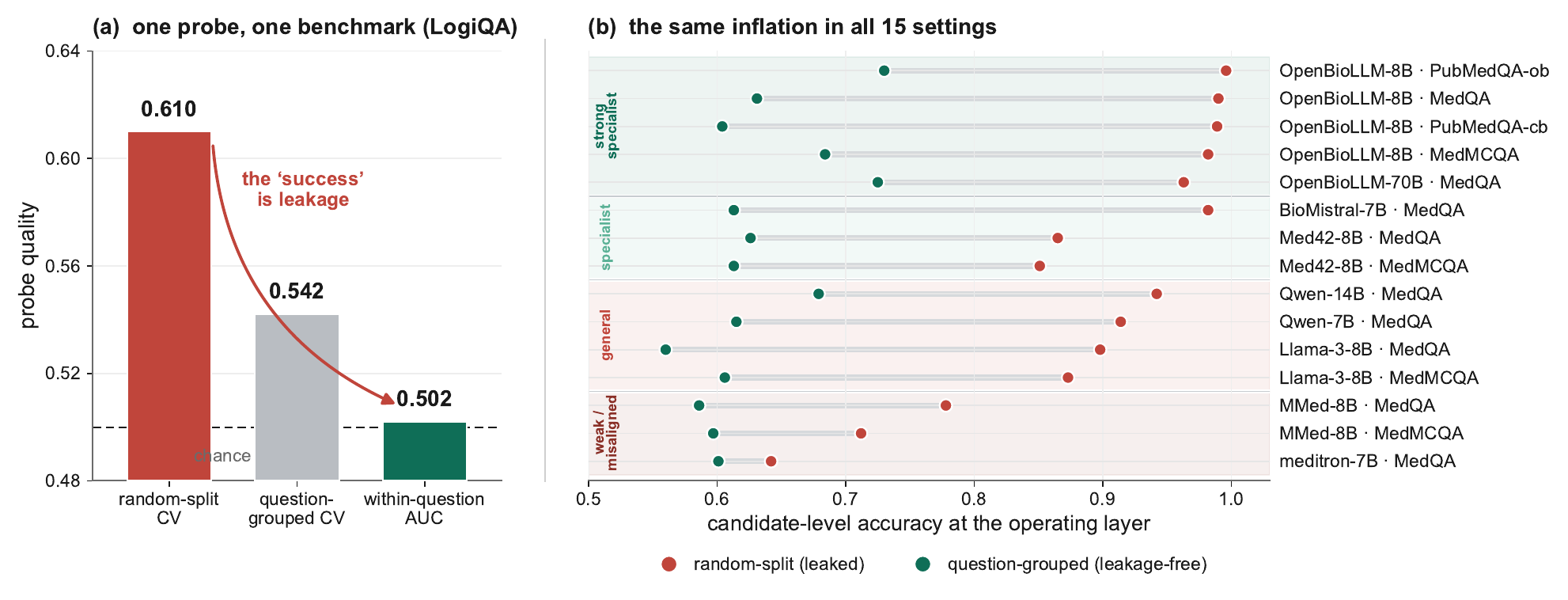}
\caption{The naive probe's apparent success is question-identity leakage. \textbf{(a)}~On LogiQA (Qwen2.5-7B, 651 questions), the naive random-split cross-validation looks informative ($0.610$), but grouping folds by question drops it to $0.542$ and the selection-relevant within-question ranking AUC is at chance ($0.502$). \textbf{(b)}~The leakage is not an artifact of that single cell: at each setting's answer-token operating layer, random-split candidate accuracy (leaked) exceeds the leakage-free question-grouped accuracy in all 15 medical settings (Supplementary Table~S17). Exact LogiQA values are in the text.}
\label{fig:leak}
\end{figure}

The natural implementation---a linear probe on a late-layer hidden state, validated by ordinary cross-validation---appears to select well, but this apparent strength is a leakage artifact. On Qwen2.5-7B over all 651 LogiQA questions (penultimate layer, mean over final tokens, random-split cross-validation), the single agent scores 51.5\%, majority vote 52.7\%, and CASE 51.7\%---a $+0.2$-point change over the single agent ($p=0.66$). Random-split 5-fold cross-validation reports 0.610 accuracy, but grouping the folds by question drops this to 0.542, and the within-question ranking AUC---what selection actually requires---is 0.502, i.e.\ chance (Fig.~\ref{fig:leak}a). Random splits encourage the probe to identify the question (whose candidates are predominantly correct or predominantly incorrect) rather than judge candidate correctness; that identity signal cannot help choose among a single question's candidates. This leakage has a computable ceiling: Proposition~4's pure-question-identity bound $\mathbb{E}_q[\max(\pi_q,1-\pi_q)]$ evaluates to $\approx0.87$ on this pool, so the inflated $0.610$ sits between leakage-free chance ($0.502$) and this identity ceiling---consistent with a random-split probe that recovers question identity only partially, since a question's candidates are divided across folds. We therefore adopt question-grouped evaluation throughout and, finding the penultimate-layer mean uninformative, switch to the answer-token representation at a late layer (Section~\ref{sec:method}). This leakage is not specific to LogiQA or to the penultimate/mean-token readout that this single-cell demonstration also happens to use: at each setting's answer-token operating layer, random-split cross-validation still inflates candidate accuracy over question-grouped evaluation in all 15 medical settings (by $+0.04$ to $+0.39$, median $\approx+0.27$), and a label-permutation null confirms the grouped estimate is leakage-free (it sits at chance in every setting; Supplementary Table~S17).

\subsection{Internal fusion beats voting only for knowledge-aligned medical models}\label{sec:when}
\begin{figure}[t]\centering
\includegraphics[width=0.92\linewidth,
trim={0bp 5bp 0bp 0bp},
clip]{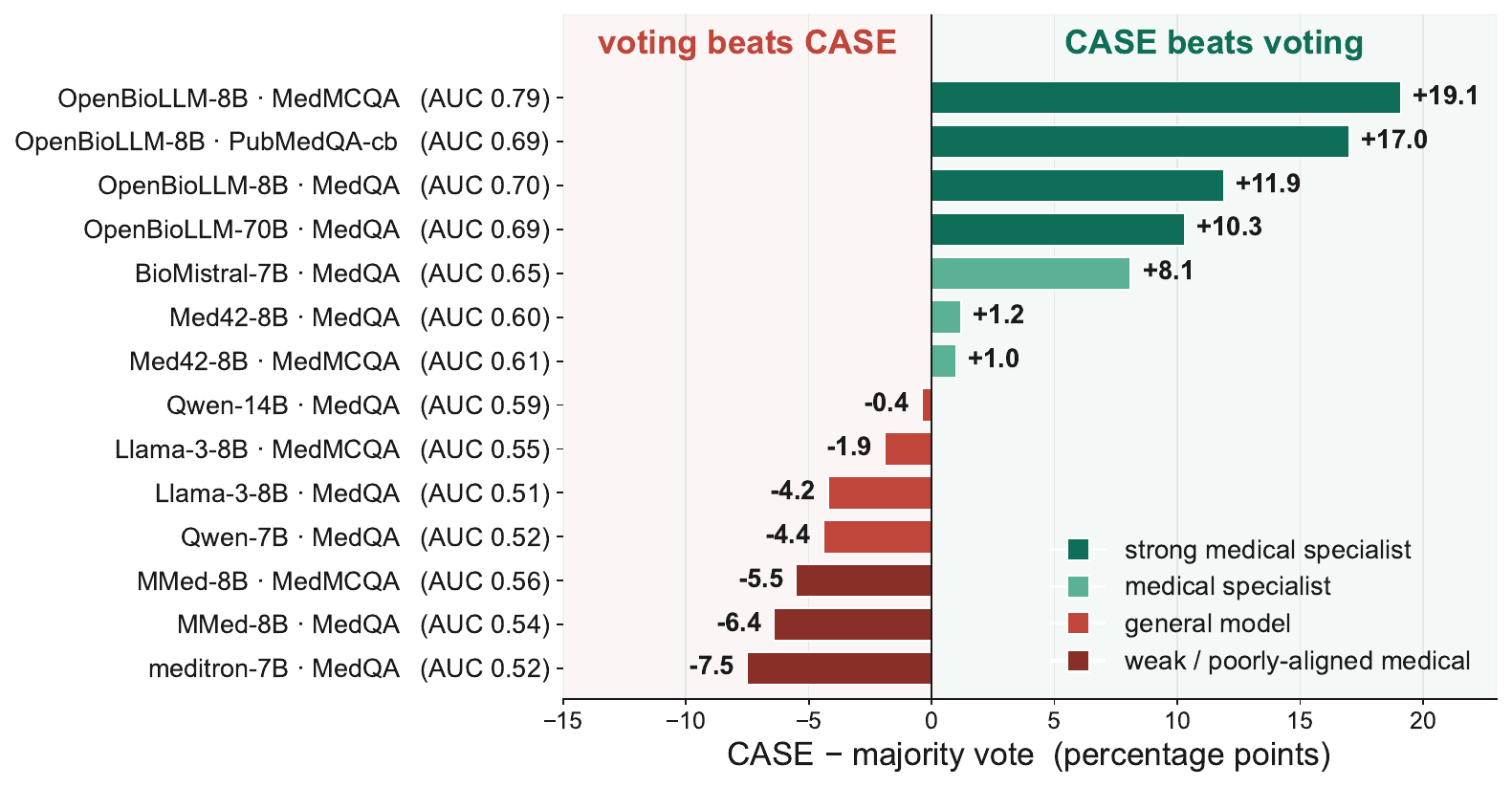}
\caption{CASE minus majority vote (percentage points, medium difficulty) across models and benchmarks. Only knowledge-aligned medical models exceed voting; general and weakly-aligned medical models fall below it. Each row's within-question decodability (AUC) is shown in parentheses: the gain crosses zero exactly as the AUC crosses the $\approx0.60$ threshold, so the decodability law can be read row by row.}
\label{fig:core}
\end{figure}

Across models and benchmarks, internal fusion outperforms voting mainly for strongly aligned medical models (Fig.~\ref{fig:core}; Table~\ref{tab:main}; the full medium breakdown for all 15 settings, with hard-bin rows for the archived-pool settings, is Supplementary Table~S1). The clearest case is OpenBioLLM-8B, which exceeds voting by $+19.1$ points on MedMCQA (95\% CI $[+11.1, +27.0]$, $p<0.001$), $+17.0$ points on closed-book PubMedQA ($[+7.3, +26.3]$, $p=0.001$), and $+11.9$ points on MedQA ($[+0.6, +23.5]$, $p=0.039$); its 70B counterpart gives $+10.3$ ($p=0.15$ at medium) and BioMistral-7B $+8.1$ ($p=0.07$), both positive but not significant at medium difficulty given their small medium bins. Med42-8B is positive but small ($+1.0$ to $+1.2$) and not significant. In contrast, internal fusion underperforms voting for general models (Qwen-7B $-4.4$, Qwen-14B $-0.4$; Llama-3-8B $-4.2$ on MedQA, $-1.9$ on MedMCQA) and for the weakly aligned medical models (meditron-7B $-7.5$ and MMed-Llama-3-8B $-5.5$ to $-6.4$); we show in Section~\ref{sec:cost} that these general models can nonetheless succeed on domains where their latent knowledge runs deep, such as graduate physics. Medium-difficulty significance is thus concentrated in OpenBioLLM; the broader evidence for the strong models comes from the hard bin, where OpenBioLLM mitigates voting's \emph{Byzantine} failure across all three benchmarks with strong statistical significance ($p<0.001$; Table~\ref{tab:main}). That the largest medium-difficulty wins come from one family is a real limitation of this medical panel---among the tested specialists only OpenBioLLM is strongly enough aligned to clear the threshold. The diagnostic, not any single model, is what generalizes: a general model clears the threshold and gains on graduate physics (Section~\ref{sec:cost}), and the signal is carried by the internal states rather than the calibration labels (a text classifier on the same labels is at chance, Section~\ref{sec:cheap})---so decodability, not the ``OpenBioLLM'' label, is what predicts the gain. Thus, the label ``medical specialist'' is not sufficient on its own: meditron is a medical model but still fails. The next subsections identify the conditions under which the internal signal becomes useful.

\begin{table}[t]\centering\footnotesize
\caption{CASE vs.\ majority vote across models, benchmarks and difficulty (leakage-free; 95\% bootstrap CI; two-sided $p$). AUC is the within-question correctness ROC-AUC at the operating layer. cb $=$ closed-book; ob $=$ open-book. CASE$-$vote is computed from unrounded accuracies and can differ from the displayed vote\% and CASE\% by $\pm0.1$ pp.}
\label{tab:main}
\resizebox{\linewidth}{!}{%
\begin{tabular}{lllrrrrrrrr}
\toprule
Model & Type & Bench & Diff & $n_q$ & AUC & vote\% & CASE\% & CASE$-$vote & 95\% CI & $p$ \\ \midrule
OpenBioLLM-8B & strong & MedQA & med & 48 & 0.702 & 51.3 & 63.2 & $+11.9$ & $[+0.6,+23.5]$ & .039 \\
OpenBioLLM-8B & strong & MedQA & hard & 288 & 0.606 & 3.5 & 11.1 & $+7.5$ & $[+5.3,+9.8]$ & $<$.001 \\
OpenBioLLM-8B & strong & MedMCQA & med & 73 & 0.794 & 50.6 & 69.7 & $+19.1$ & $[+11.1,+27.0]$ & $<$.001 \\
OpenBioLLM-8B & strong & MedMCQA & hard & 296 & 0.706 & 3.9 & 13.2 & $+9.3$ & $[+7.1,+11.6]$ & $<$.001 \\
OpenBioLLM-8B & strong & PubMedQA-cb & med & 71 & 0.685 & 48.3 & 65.3 & $+17.0$ & $[+7.3,+26.3]$ & .001 \\
OpenBioLLM-8B & strong & PubMedQA-cb & hard & 286 & 0.694 & 4.5 & 17.9 & $+13.4$ & $[+10.8,+16.4]$ & $<$.001 \\
OpenBioLLM-8B & strong & PubMedQA-ob & med & 29 & 0.650 & 53.2 & 61.9 & $+8.7$ & $[-6.8,+22.7]$ & .259 \\
OpenBioLLM-70B & strong & MedQA & med & 21 & 0.691 & 49.5 & 59.8 & $+10.3$ & $[-4.4,+24.8]$ & .153 \\
Med42-8B & spec & MedQA & med & 48 & 0.605 & 53.8 & 55.0 & $+1.2$ & $[-6.2,+8.9]$ & .761 \\
Med42-8B & spec & MedMCQA & med & 39 & 0.610 & 54.2 & 55.2 & $+1.0$ & $[-6.8,+9.4]$ & .843 \\
BioMistral-7B & spec & MedQA & med & 86 & 0.645 & 47.1 & 55.2 & $+8.1$ & $[-0.4,+16.3]$ & .071 \\
meditron-7B & weak & MedQA & med & 153 & 0.518 & 54.6 & 47.1 & $-7.5$ & $[-10.5,-4.7]$ & $<$.001 \\
MMed-8B & weak & MedQA & med & 58 & 0.543 & 54.8 & 48.4 & $-6.4$ & $[-12.7,-0.0]$ & .050 \\
MMed-8B & weak & MedMCQA & med & 84 & 0.555 & 56.2 & 50.8 & $-5.5$ & $[-10.4,-0.6]$ & .033 \\
Qwen2.5-7B & general & MedQA & med & 28 & 0.516 & 51.5 & 47.1 & $-4.4$ & $[-14.0,+4.4]$ & .341 \\
Qwen2.5-14B & general & MedQA & med & 18 & 0.592 & 55.6 & 55.3 & $-0.4$ & $[-14.1,+13.1]$ & .959 \\
Llama-3-8B & general & MedQA & med & 49 & 0.513 & 54.5 & 50.3 & $-4.2$ & $[-12.9,+4.3]$ & .355 \\
Llama-3-8B & general & MedMCQA & med & 64 & 0.547 & 50.6 & 48.8 & $-1.9$ & $[-8.6,+4.9]$ & .599 \\
\bottomrule
\end{tabular}}
\end{table}

To make the mechanism concrete, Table~\ref{tab:worked} shows three real cases from this regime (OpenBioLLM, MedQA): the correct answer is sampled by only a minority of candidates, so majority voting selects the wrong option, whereas the gate assigns the highest internal correctness score to the rare correct candidate and recovers it.

\begin{table}[t]\centering\footnotesize
\caption{Worked examples (OpenBioLLM-8B, MedQA; $N=16$ sampled candidates; leakage-free question-grouped gate). For each item: the option votes and mean CASE gate $g$, the majority-vote pick (wrong, \textcolor{loss}{red}), and the CASE pick (correct gold, \textcolor{winteal}{teal}).}
\label{tab:worked}
\begin{tabular}{p{5.0cm}p{4.6cm}cc}
\toprule
Question (abridged) & Options --- votes $\cdot$ gate $g$ & Vote & CASE \\ \midrule
Man with an itchy annular abdominal plaque; KOH prep confirms hyphae. Next best step in management? &
A. Itraconazole --- 6$\times\cdot$g\,0.90 \textcolor{winteal}{(gold)}\newline B. Griseofulvin --- 10$\times\cdot$g\,0.31\newline C. Topical clindamycin --- 0$\times$\newline D. Doxycycline --- 0$\times$ &
\textcolor{loss}{\textbf{B}} & \textcolor{winteal}{\textbf{A}} \\ \addlinespace
Boy with seizure disorder, apical murmur, calcified retinal lesions. Mutation in which gene? &
A. NF1 (chr 17) --- 8$\times\cdot$g\,0.71\newline B. NF2 (chr 22) --- 0$\times$\newline C. TSC1 (chr 9) --- 8$\times\cdot$g\,1.00 \textcolor{winteal}{(gold)}\newline D. VHL (chr 3) --- 0$\times$ &
\textcolor{loss}{\textbf{A}} & \textcolor{winteal}{\textbf{C}} \\ \addlinespace
Drug inhibits phosphate release by the myosin head; which cross-bridge step is blocked? &
A. Myosin head cocking --- 12$\times\cdot$g\,0.00\newline B. Exposure of binding sites --- 0$\times$\newline C. Myosin binding to actin --- 0$\times$\newline D. Power stroke --- 4$\times\cdot$g\,0.29 \textcolor{winteal}{(gold)} &
\textcolor{loss}{\textbf{A}} & \textcolor{winteal}{\textbf{D}} \\
\bottomrule
\end{tabular}
\end{table}

\subsection{Single-variable controls isolate aligned latent knowledge}\label{sec:controls}
\begin{figure}[t]\centering
\includegraphics[width=\linewidth,
trim={0bp 12bp 0bp 0bp},
clip]{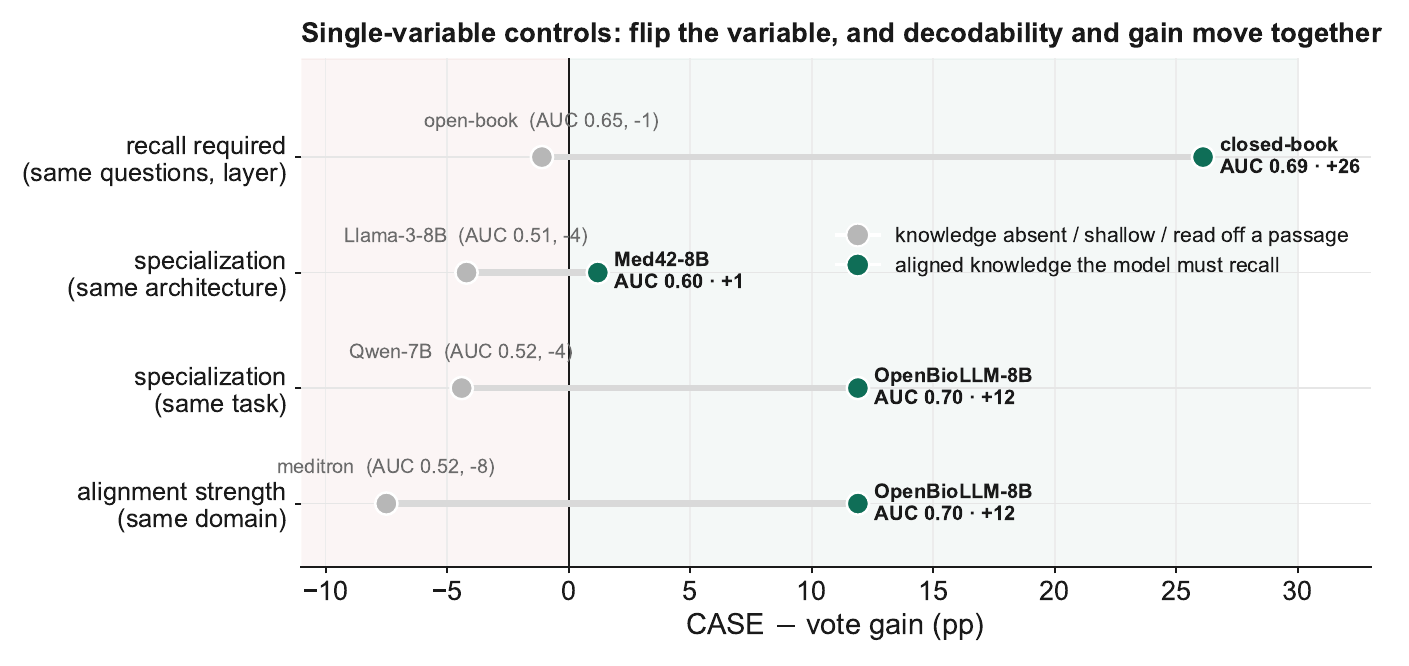}
\caption{Single-variable controls. Each row changes one variable and holds the rest fixed; in every case the condition that requires the model to recall aligned knowledge it holds (teal) both raises the within-question decodability (AUC) and moves the fusion gain from below voting to above it, while the matched control (grey)---knowledge absent, shallow, or readable off a supplied passage---does neither. The recall row's gains are the same-question, same-layer closed- vs.\ open-book comparison (Supplementary Table~S16); each point's AUC is that condition's own operating-layer decodability (closed-book L24, open-book L20); the others hold the architecture, the task, or the domain fixed.}
\label{fig:controls}
\end{figure}

Four single-variable controls identify the likely source of the effect and address several obvious confounds (Fig.~\ref{fig:controls}); in each, the condition requiring recall of aligned knowledge moves both the decodability and the gain, and its matched control moves neither.

\recipeitem{%
\textit{Same architecture.} On the same Llama-3-8B backbone, the general Llama-3-8B-Instruct model (MedQA accuracy 0.60) loses to voting ($-4.2$) while the medically aligned Med42-8B edges above it ($+1.2$ on the same task). Neither medium-difficulty gain is individually significant ($p=0.36$ and $0.76$), so we rest the comparison on the more stable within-question decodability, which separates the two (0.513 general vs.\ 0.605 specialist). The general model is more accurate overall, so the effect tracks specialization rather than architecture or raw task accuracy.}
\recipeitem{%
\textit{Closed- vs.\ open-book, at a fixed layer.} On one fixed question set---the 71 closed-book-medium PubMedQA questions scored in both conditions at the same late layer (gold answers align 600/600)---removing the abstract, which forces parametric recall, yields $+26.1$ over voting while providing it (open-book reading) yields only $-1.1$ ($+27.3$ vs.\ $+0.8$ at a second late layer; Fig.~\ref{fig:controls} row~1, Supplementary Table~S16). At a fixed operating layer the correctness signal is thus passage-dependent. Open-book correctness is not undecodable, however: at its own earlier operating layer the open-book representation is itself decodable and beats voting (AUC 0.650, $+8.7$; Table~\ref{tab:law}). Passage removal therefore relocates and strengthens a recall-based correctness signal at the late operating layer, rather than being the only condition under which correctness is decodable.}
\recipeitem{%
\textit{Specialist vs.\ general on the same task.} As shown in Section~\ref{sec:when}, the specialist setting gives $+11.9$ on MedQA, whereas the general setting gives $-4.4$.}

\recipeitem{%
\textit{Strong vs.\ weak specialist.} Weakly aligned medical models, including meditron (accuracy 0.27, near the floor) and MMed, show chance-level decodability across all layers and underperform voting. Thus, the boundary is not the medical label itself, but the presence of reliable, non-floor domain knowledge.}

Together these controls suggest that the late-layer fusion signal reflects recall of aligned domain knowledge: it is present when a well-aligned specialist must retrieve the answer internally, and at that operating layer it is absent when the answer can instead be read from a supplied passage---though, as the open-book point shows, correctness can still be decodable at another layer.

\subsection{A single law predicts when internal fusion helps}\label{sec:law}
\begin{figure}[t]\centering
\includegraphics[width=\linewidth,
trim={0bp 30bp 0bp 0bp},
clip
]{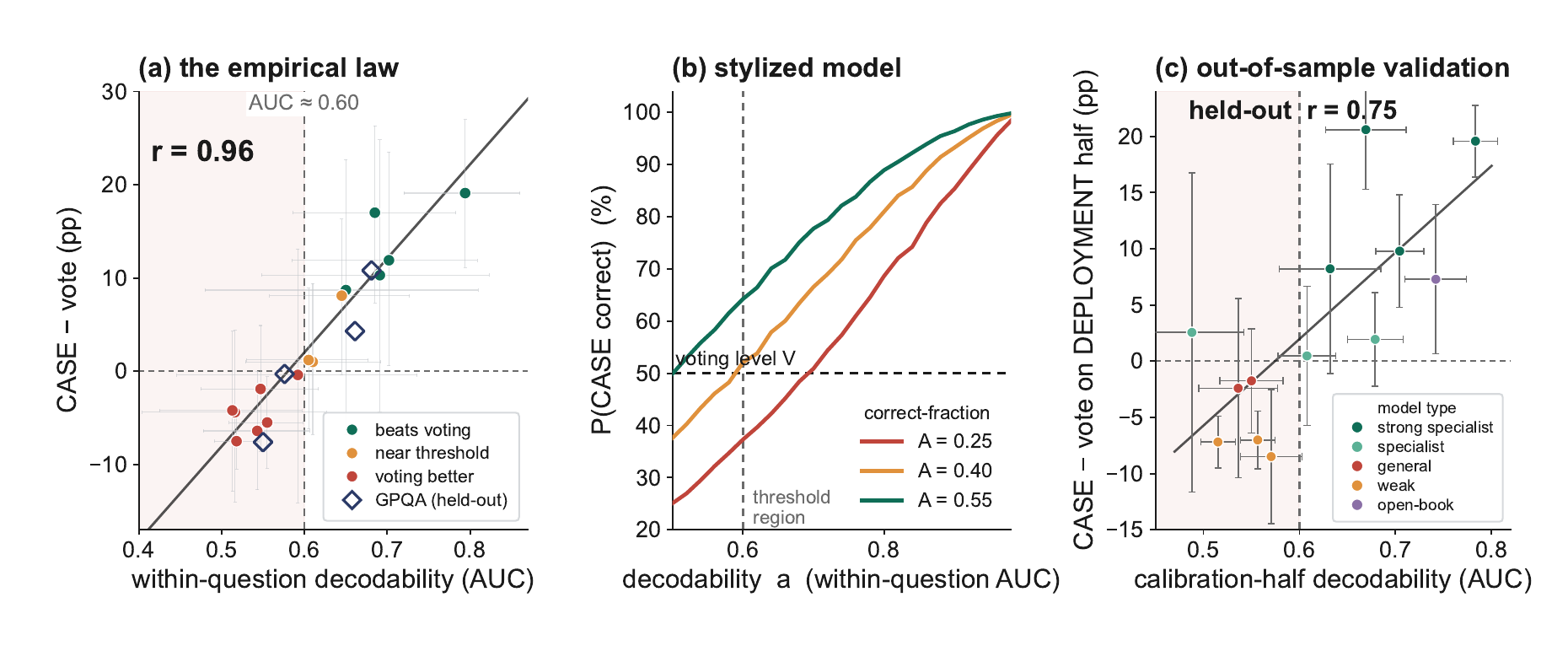}
\caption{Within-question decodability predicts the fusion gain. \textbf{(a)}~The empirical law: across 15 medical settings (filled, colored by decodability region) the within-question AUC predicts the CASE-minus-voting gain (Pearson $r=0.96$; fitted line), with a clean threshold at $\mathrm{AUC}\approx0.60$; open diamonds are out-of-sample GPQA settings (Section~\ref{sec:cost}), not used in the fit. Error bars are 95\% confidence intervals (horizontal: question-bootstrap of the within-question AUC; vertical: bootstrap of the gain); the AUC is markedly tighter than the gain, consistent with it being the more stable indicator. \textbf{(b)}~The stylized model of Section~\ref{sec:analysis}: predicted selection accuracy rises with decodability from the pool accuracy $A$ toward $1$ (Proposition~1) and overtakes the voting level $V$ at a threshold above $\tfrac12$ (Proposition~2); the three curves are three correct-fractions $A$. \textbf{(c)}~Out-of-sample validation: decodability measured on a calibration half predicts the deployment-half gain on held-out questions ($r=0.75$; 13 settings, 10 splits each, mean $\pm$ s.d.).}
\label{fig:law}
\end{figure}

The heterogeneous outcomes of Sections~\ref{sec:when}--\ref{sec:controls} are explained by a single predictor, in the form the analysis of Section~\ref{sec:analysis} predicts. Across 15 model--benchmark settings, all evaluated at the pooled-best late operating layer, the leakage-free within-question correctness AUC predicts the CASE-minus-voting gain with Pearson $r=0.96$ ($p=3\times10^{-8}$; Spearman $\rho=0.92$), with a decision boundary at $\mathrm{AUC}\approx0.60$ that separates the success and failure groups with no overlap (success AUC 0.605--0.794; failure 0.513--0.592; Table~\ref{tab:law}, Fig.~\ref{fig:law}). The least-squares fit is
\begin{equation}\label{eq:fit}
\widehat{\Delta}(\AUCwq)=100.8\,\AUCwq-58.5,\qquad r=0.96,
\end{equation}
crossing zero at $\mathrm{AUC}=0.58$, inside the empirical success/failure gap. The slope has standard error $8.6$ (OLS 95\% CI $[82,119]$; a model-cluster bootstrap over the nine model clusters widens it to $[62,116]$; an errors-in-variables Deming fit that accounts for measurement error in the AUC predictor gives a similar $105.8$, so the OLS slope is not appreciably attenuated), and the gain scatters about the line with residual standard deviation $2.7$ pp, so Eq.~\eqref{eq:fit} should be read as a forecast with a prediction interval of roughly $\pm6$ pp (a leverage-aware interval widens to $\pm6.8$ pp at the extremes of the AUC range) rather than an exact point estimate. Two cautions temper this in-sample fit. First, the separating threshold is itself uncertain: the same cluster bootstrap places the zero-crossing at $0.58$ with 95\% CI $[0.562,0.596]$, a width ($0.034$) larger than the empirical success/failure gap ($0.013$), so we treat a narrow band around the boundary (roughly $\mathrm{AUC}\,0.55$--$0.65$) as indeterminate rather than as a sharp cutoff. Second, because the predictor (within-question AUC) and the outcome (the selection gain) are computed from the same gate on the same questions, part of the tight fit is structural: the residual variance is only about a third of what the gains' own sampling error would produce on its own (residual s.d.\ $2.7$ pp vs.\ sampling s.d.\ $\approx4.5$ pp), so the two quantities' errors move together. We therefore read $r=0.96$ as an internal consistency check and rest the predictive claim on the held-out calibration-to-deployment test below ($r=0.75$). Decodability is therefore an a-priori deployment diagnostic: measuring it for a given model--domain pair on a small labeled set, before committing to the fusion operator, estimates whether hidden-state selection is likely to help. It also unifies the specialist/general, closed/open-book, scale, and strong/weak-model effects under a single quantity. The relationship is quantitative and falsifiable, issuing testable point predictions: it predicts the out-of-sample GPQA physics gains to within 3.8 points from the medical fit alone (Section~\ref{sec:cost}), and it is falsifiable in that, on the leakage-free medium-bin decodability axis, any setting above the threshold that failed to gain, or below it that gained, would break it. We turn this diagnostic into an explicit recipe in Section~\ref{sec:recipe}.

Three checks indicate the law is not driven by a few influential points or by non-independence among settings that share a model. A leave-one-point-out jackknife keeps the correlation in the range $r\in[0.947, 0.973]$. A leave-one-model-out analysis---dropping all settings of a given model, nine model clusters in total---gives $r\in[0.941, 0.959]$. As the most stringent form of this test, we drop the entire OpenBioLLM family at once---all five settings of the 8B and 70B variants, a third of the panel and its five largest gains: the correlation remains $r=0.90$ ($p<0.001$, $n=10$), with the fitted slope (91 vs.\ 101 gain points per unit AUC) and the implied operating threshold (AUC 0.587 vs.\ 0.580) essentially unchanged. The law is therefore not an artifact of one model family. A block bootstrap that resamples whole model clusters, accounting for within-model dependence rather than over-counting individual settings, yields a 95\% confidence interval of $r\in[0.881, 0.983]$. The fit is also insensitive to the difficulty-bin edges: recomputing the law under six alternative medium-bin definitions (from $0.30$--$0.65$ to $0.40$--$0.55$) keeps $r\in[0.90, 0.97]$ (Supplementary Table~S2). The success/failure separation is complete though the margin is narrow: the highest within-question AUC among the failing settings is 0.592 and the lowest among the succeeding settings is 0.605, so any threshold in this gap (we use its midpoint, 0.60) classifies all 15 settings correctly, with the boundary settings themselves near zero gain (Supplementary Table~S2).

As a further check, the diagnostic is not driven by choosing the operating layer on the same data used for scoring. We repeated the analysis with held-out layer selection. For each of three random splits, the operating late layer was selected on one half of the questions, and the within-question AUC and medium-difficulty selection gain were measured on the other, disjoint half. We ran this for four representative above-threshold settings---the three OpenBioLLM settings and BioMistral---each with a 600-candidate pool. For the three OpenBioLLM settings the held-out within-question AUC remains above the threshold (MedQA $0.715\pm0.014$, MedMCQA $0.786\pm0.025$, closed-book PubMedQA $0.654\pm0.018$; mean $\pm$ sd over splits) and the held-out selection gain remains clearly positive ($+4.8\pm3.2$, $+13.5\pm5.6$, $+14.5\pm6.8$ pp; Supplementary Table~S3), so the operating point is not an artifact of choosing the layer on the scored data. The gains attenuate from their in-sample values mainly because halving leaves a small medium-difficulty subsample; the AUC, computed over all mixed-correctness questions, is the more stable indicator and does not. The borderline BioMistral setting (in-sample $+8.1$, $p=0.07$) instead falls to a below-threshold held-out AUC ($0.555\pm0.019$) and a near-zero held-out gain ($+0.2\pm1.8$ pp): so close to the threshold, halving the data both weakens the gate and destabilizes the layer choice---the fragility the law anticipates near the boundary. The check therefore confirms the operating point where decodability is high and reproduces the expected marginality where it is low.

Three further audits concern the operating point. First, the three-seed regenerations of the two strongest settings, scored on the law's own medium-bin AUC axis (rather than a pooled AUC), fall within its $\pm6$ pp prediction interval---for example MedQA seed~2, AUC $0.810$, predicts $+23$ and gains $+22.2$ (Supplementary Table~S4); the seed spread is thus an independent validation of the law, not a perturbation of it. Second, we audited the late-half layer restriction on the settings that define the success boundary: for Med42 the unconstrained per-layer argmax lies inside the late half on MedQA but one layer below the boundary on MedMCQA (L14, pooled AUC $0.668$ vs.\ the constrained L16, $0.641$; the medium operating-point value in Table~\ref{tab:law} is $0.610$), so the reported operating-layer AUC is a mild lower bound under the constraint, but the difference does not change any setting's classification. Third, we extended the held-out layer check to the two settings on the success boundary (Med42-8B and Qwen2.5-14B) and to a below-threshold \emph{failing} setting (meditron-MedQA): under held-out layer selection the boundary settings reproduce or exceed their operating-layer AUC (Med42-MedMCQA $0.612\pm0.013$, Med42-MedQA $0.669\pm0.022$, Qwen-14B $0.621\pm0.011$; Supplementary Table~S3), while the failing setting stays at chance ($0.503\pm0.012$), so a sub-threshold AUC is not itself inflated by the layer search and the success/failure ordering is not an artifact of selection. Finally, a within-question label-permutation null validates the layer search itself: permuting correctness labels within each question and repeating the entire late-layer argmax gives a best-layer AUC near chance (mean $0.51$--$0.53$), so the operating-layer values ($0.63$--$0.73$ for the above-threshold and boundary settings) exceed the null layer-search by $0.11$--$0.22$ AUC (permutation $p=0.01$ for the tightest boundary case, Med42-MedMCQA; Supplementary Table~S20), whereas the failing meditron-MedQA setting sits only $0.015$ above its null, at chance.

Because the predictor and the outcome are read from the same gate scores, we also verify the relationship out of sample. In a calibration-to-deployment test---decodability measured on one random half of each setting's questions, and the CASE-minus-voting gain measured on the disjoint other half with a gate trained only on the calibration half---calibration decodability still predicts deployment gain at Pearson $r=0.75$ (bootstrap 95\% CI $[0.51,0.93]$; $p=3\times10^{-3}$; 13 settings, 10 splits each), and the $\mathrm{AUC}\approx0.60$ threshold separates 11 of 13 settings, the two exceptions near zero (Supplementary Fig.~S1). Because the in-sample law (Table~\ref{tab:law}) and this held-out test both use the same pooled-best operating-layer recipe---the layer here chosen on the calibration half alone---this is a like-for-like out-of-sample validation: the relationship is predictive, not an artifact of the shared scores. (Two general-model settings are too small to split.) Out of sample the forecast is looser than the in-sample fit, as expected: the medical law's point prediction covers the held-out deployment gain within its $\pm6$ pp band for 8 of 13 settings (RMSE $6.4$ pp), so as a deployment forecast it should be read with a wider interval, roughly $\pm12$ pp. The held-out correlation's spread is carried mainly by the settings that straddle the threshold; the below-threshold cluster is predicted to gain little and does, so the operative claim---that the diagnostic separates deployable from non-deployable settings---is more robust than the coefficient alone, and its \emph{generality} rests on the prospective cross-domain test of Section~\ref{sec:cost} rather than on within-medical resampling.

\begin{table}[t]\centering\footnotesize
\caption{The 15 model--benchmark points underlying the decodability$\rightarrow$gain law, all at the pooled-best late operating layer (held-out $r=0.75$, Supplementary Fig.~S1; in-sample Pearson $r=0.956$ as a consistency check, $p=3\times10^{-8}$, Spearman $\rho=0.92$; separating threshold $\mathrm{AUC}\approx0.60$). The \emph{medium} column is the medium-difficulty gain that the law predicts (the mechanistic result); \emph{uncond.} is the difficulty-unconditional gain over the whole candidate pool---the number a deployer who cannot bin questions by difficulty would experience. The unconditional gain is roughly a third of the medium-bin gain (easy questions dominate the pool and CASE does not help there), but near the boundary it scatters within $\pm0.4$ pp of zero (e.g.\ Med42-8B MedMCQA $-0.1$ just above threshold, meditron-7B $+0.2$ just below) and is clearly positive only well above threshold, so the diagnostic predicts the sign of the deployable gain away from the boundary.}
\label{tab:law}
\begin{tabular}{lrrrrr}
\toprule
Model / benchmark & within-Q AUC & vote \% & CASE \% & medium (pp) & uncond.\ (pp) \\ \midrule
OpenBioLLM-8B MedMCQA & 0.794 & 50.6 & 69.7 & $+19.1$ & $+6.8$ \\
OpenBioLLM-8B MedQA & 0.702 & 51.3 & 63.2 & $+11.9$ & $+4.0$ \\
OpenBioLLM-70B MedQA & 0.691 & 49.5 & 59.8 & $+10.3$ & $+0.7$ \\
OpenBioLLM-8B PubMedQA-cb & 0.685 & 48.3 & 65.3 & $+17.0$ & $+6.0$ \\
OpenBioLLM-8B PubMedQA-ob & 0.650 & 53.2 & 61.9 & $+8.7$ & $+1.7$ \\
BioMistral-7B MedQA & 0.645 & 47.1 & 55.2 & $+8.1$ & $+2.7$ \\
Med42-8B MedMCQA & 0.610 & 54.2 & 55.2 & $+1.0$ & $-0.1$ \\
Med42-8B MedQA & 0.605 & 53.8 & 55.0 & $+1.2$ & $+1.6$ \\
Qwen-14B MedQA & 0.592 & 55.6 & 55.3 & $-0.4$ & $+0.3$ \\
MMed-8B MedMCQA & 0.555 & 56.2 & 50.8 & $-5.5$ & $-1.0$ \\
Llama-3-8B MedMCQA & 0.547 & 50.6 & 48.8 & $-1.9$ & $+0.4$ \\
MMed-8B MedQA & 0.543 & 54.8 & 48.4 & $-6.4$ & $-1.4$ \\
meditron-7B MedQA & 0.518 & 54.6 & 47.1 & $-7.5$ & $+0.2$ \\
Qwen-7B MedQA & 0.516 & 51.5 & 47.1 & $-4.4$ & $-1.9$ \\
Llama-3-8B MedQA & 0.513 & 54.5 & 50.3 & $-4.2$ & $-1.1$ \\
\bottomrule
\end{tabular}
\end{table}

\subsection{In the Byzantine regime, fusion improves with $N$ while voting collapses}\label{sec:nscale}
\begin{figure}[t]\centering
\includegraphics[width=\linewidth,
trim={0bp 10bp 0bp 0bp},
clip]{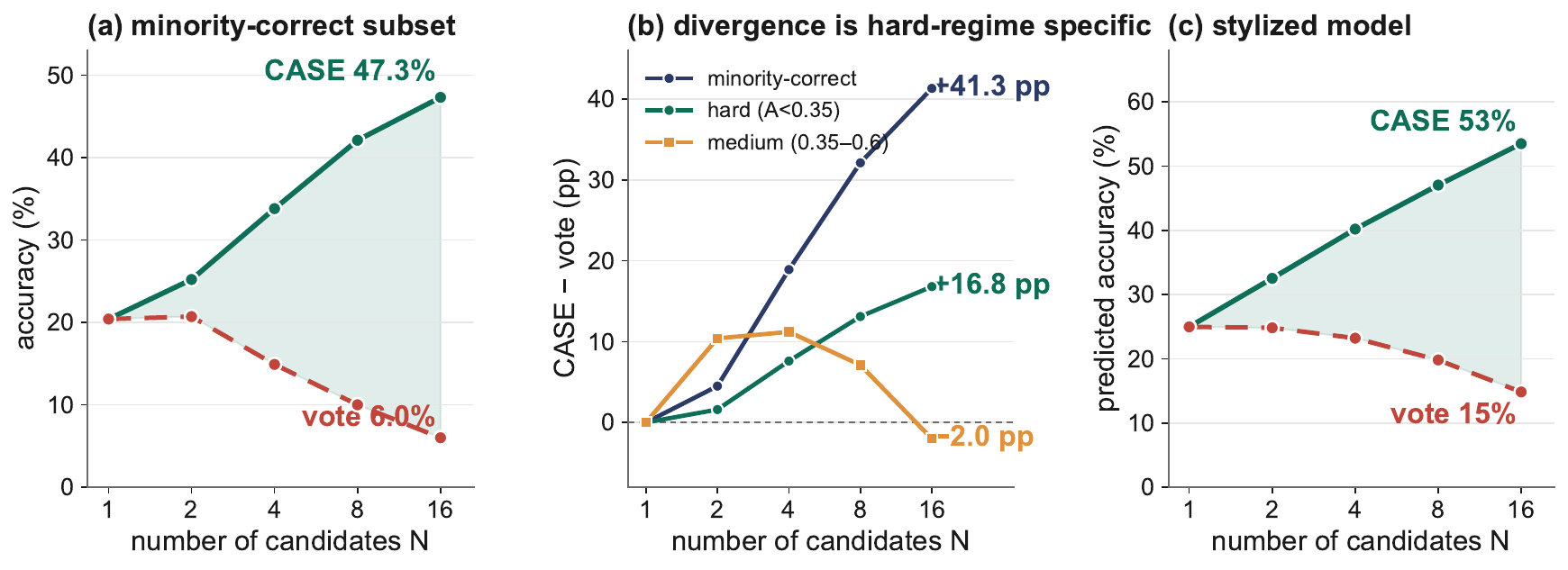}
\caption{\emph{Byzantine} $N$-scaling. \textbf{(a)}~On the minority-correct subset, as the candidate count $N$ grows CASE climbs from $20.4$ to $47.3\%$ while majority voting collapses from $20.4$ to $6.0\%$. \textbf{(b)}~Resolved by difficulty, the CASE-minus-vote gain diverges on the minority-correct and hard bins ($+41.3$ and $+16.8$ pp at $N=16$) but reverses on the medium bin ($-2.0$), where voting recovers---the advantage is specific to the \emph{Byzantine} hard regime. \textbf{(c)}~The stylized model (Proposition~3, $a=0.70$) reproduces the divergence: predicted CASE accuracy rises with $N$ while voting falls.}
\label{fig:nscale}
\end{figure}

On the subset where the correct answer is a minority (OpenBioLLM, MedQA; 131 questions), the two fusion rules diverge as the candidate count $N$ increases (Table~\ref{tab:nscale}, Fig.~\ref{fig:nscale}). Majority-vote accuracy decreases from 20.4\% to 6.0\% as $N$ grows from 1 to 16, because voting increasingly selects the wrong majority answer. In contrast, CASE increases from 20.4\% to 47.3\%, because additional candidates provide more opportunities to identify the rare correct one. The advantage reaches $+41.3$ points at $N=16$. Because this subset is defined using ground truth, it is a diagnostic rather than something a deployer can target directly; the same divergence holds on the broader hard bin (288 questions), a difficulty stratum rather than a hand-picked label pattern---though it, too, is gold-defined, so the deployable figure is the unconditional gain of Section~\ref{sec:recipe}, not the per-bin one: as $N$ grows to 16 majority accuracy collapses from 6.2\% to 0.2\% while CASE rises to 17.0\%, a $+16.8$-point advantage (Supplementary Table~S11). On the medium bin the correct answer is closer to a plurality, so voting instead recovers with $N$ and overtakes CASE by $N=16$; the \emph{Byzantine} $N$-scaling advantage is thus specific to the hard regime. Consistently, across benchmarks, the selection gain is concentrated in the medium/hard \emph{Byzantine} regime and largely disappears on easy questions, where voting is already strong (Fig.~\ref{fig:anatomy}b). This is the behavior expected from hidden-state selection where the internal signal is available: it converts additional candidates into improved reliability, whereas voting can convert the same candidates into stronger wrong consensus.

\begin{table}[t]\centering\footnotesize
\caption{\emph{Byzantine} $N$-scaling (OpenBioLLM $\times$ MedQA, minority-correct subset, 131 questions). Real draws without replacement from the candidate pool. CASE$-$maj is computed from unrounded accuracies and can differ from the displayed columns by $\pm0.1$ pp.}
\label{tab:nscale}
\begin{tabular}{rrrrr}
\toprule
$N$ & single\% & majority\% & CASE\% & CASE$-$maj \\ \midrule
1 & 20.4 & 20.4 & 20.4 & $+0.0$ \\
2 & 20.7 & 20.7 & 25.2 & $+4.5$ \\
4 & 20.4 & 14.9 & 33.8 & $+18.9$ \\
8 & 20.7 & 10.0 & 42.1 & $+32.1$ \\
16 & 20.8 & 6.0 & 47.3 & $+41.3$ \\
\bottomrule
\end{tabular}
\end{table}

\subsection{A shared, transferable correctness axis: usable for selection, not steering}\label{sec:mech}
\begin{figure}[t]\centering
\includegraphics[width=0.78\linewidth,
trim={0bp 10bp 0bp 0bp},
clip
]{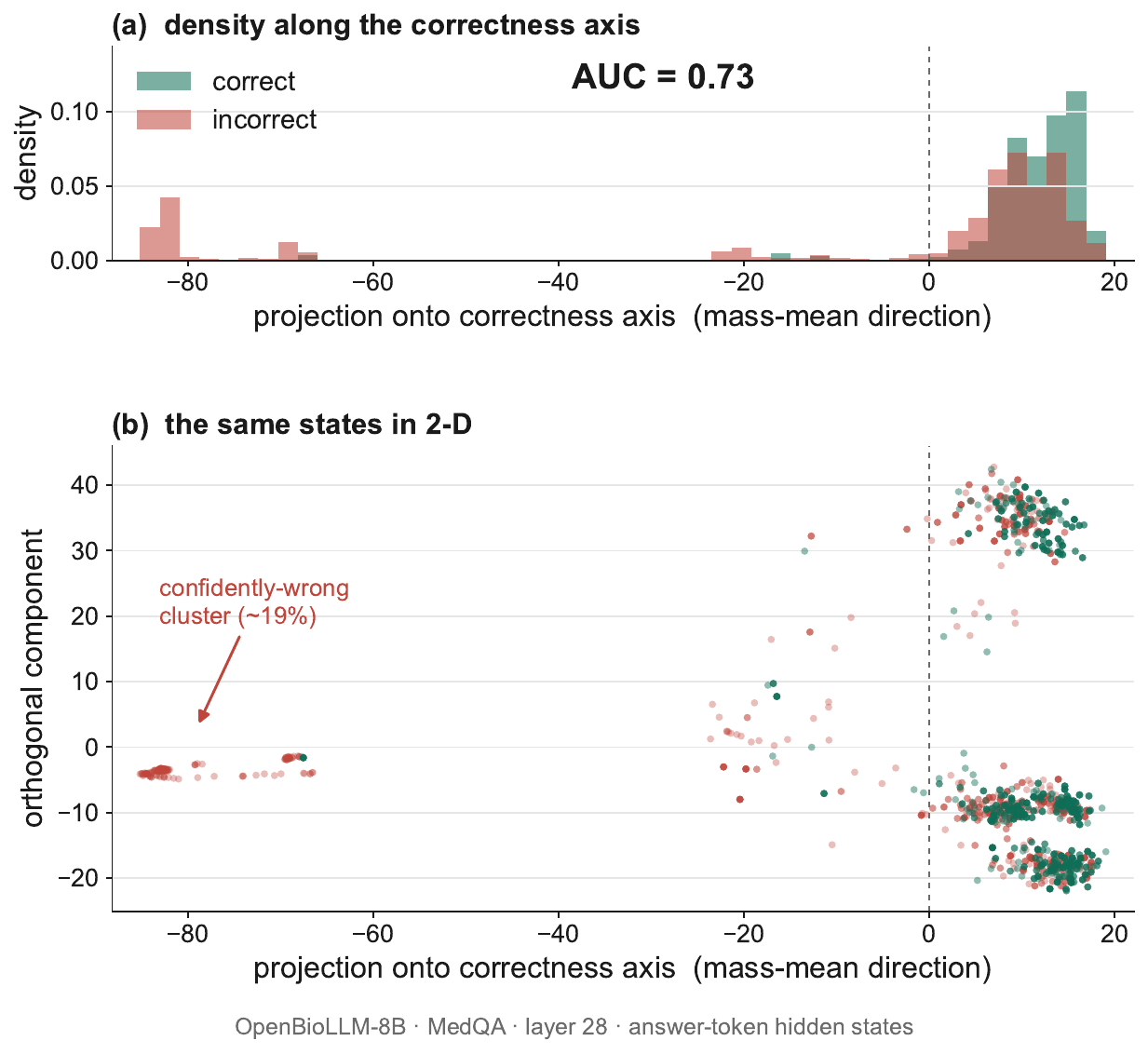}
\caption{Answer-token hidden states (OpenBioLLM-8B, MedQA, layer 28) projected onto the correctness axis (mass-mean direction). (a) Marginal density of correct (green) vs.\ incorrect (red) answers; (b) the same states in 2-D, where a $\sim$19\% cluster of incorrect answers separates cleanly on the far incorrect side---candidates the gate confidently and correctly scores as wrong, a recognizable error mode---while the harder remaining errors overlap the correct mass on the right. The pooled single-axis projection separates the two classes at AUC $=0.73$; this is the pooled projection AUC at the fixed layer 28, distinct from the within-question operating-layer decodability AUC (0.702) reported in Tables~\ref{tab:main} and \ref{tab:law}.}
\label{fig:sep}
\end{figure}

\begin{figure}[t]\centering
\includegraphics[width=\linewidth]{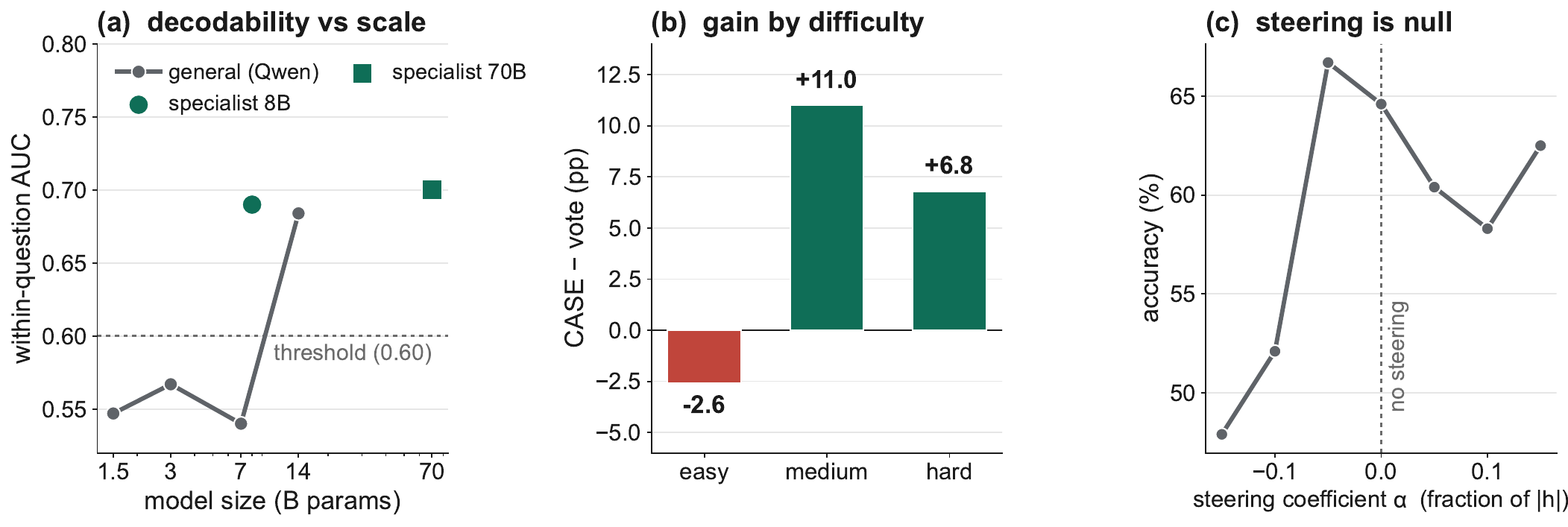}
\caption{Anatomy of the internal signal. (a) Within-question decodability emerges with scale for general models and is saturated by medical specialization. (b) The selection gain concentrates in the \emph{Byzantine} (medium/hard) regime. (c) Adding the correctness direction during generation does not improve accuracy---the readout is not a causal lever.}
\label{fig:anatomy}
\end{figure}

The correctness signal lives on a shared, transferable axis. The correctness direction---the difference of class-mean answer-token activations---is nearly identical across medical benchmarks (cosine 0.934 between MedQA and MedMCQA). A gate trained on one benchmark, moreover, selects on another at within-question AUC 0.77 (0.72 in the reverse direction), comparable to within-domain performance (Supplementary Table~S8), and the same axis transfers across physics, chemistry, and medicine (Supplementary Table~S15). This cross-benchmark and cross-domain transfer points to a shared internal representation rather than a benchmark-specific shortcut or an answer-format cue such as the chosen option letter: a format cue tied to one benchmark's answer distribution could not transfer this way. A direct within-setting control agrees: on OpenBioLLM/MedQA, once the $11.5\%$ non-parseable generations are excluded, an option-letter baseline falls to chance (within-question AUC $0.48$) while the hidden-state gate stays above it ($0.60$) and CASE still beats voting by $+7.0$ pp, so the signal is genuine correctness rather than a format prior or mere detection of malformed outputs (Supplementary Section~S13, Table~S18). When projected onto this single axis, correct and incorrect answer-token states form visibly separated distributions (Fig.~\ref{fig:sep}). This direction is a readout, not a causal lever. Across four steering variants (weak fixed-norm, strong norm-scaled, multi-layer, and a medium-difficulty-targeted sweep), adding the correctness direction to the residual stream during generation does not raise accuracy---it slightly lowers it (Fig.~\ref{fig:anatomy}c; a representative steering-coefficient sweep is Supplementary Table~S10)---and large coefficients collapse generation. The dissociation is two-sided. The direction is not a lever that \emph{writes} correctness in (adding it does not steer), but it is \emph{necessary} for reading correctness out: projecting the single mass-mean correctness direction out of the answer-token states collapses the gate from a pooled projection AUC of $0.73$ to well below chance ($0.13$ on OpenBioLLM/MedQA), so that one direction carries essentially all of the decodable correctness signal. Thus, decodability does not imply causal steerability, in contrast to the truthfulness directions used for inference-time intervention and representation engineering \cite{r7}: the signal is a genuine readout---sufficient and necessary for selection, but not a causal lever for generation.

Two further properties clarify where the signal comes from and connect it to the law (Fig.~\ref{fig:anatomy}a). First, for general-purpose models the within-question decodability emerges with scale: on LogiQA the pooled answer-token AUC rises overall from 0.55 at 1.5B to 0.68 at 14B---not monotonically, with a dip at 7B (Supplementary Table~S7)---so a small general model sits below the threshold while a large one can approach it. Medical specialization, by contrast, saturates decodability at 8B (OpenBioLLM-8B and -70B give essentially the same AUC, 0.70 vs.\ 0.69), so specialization, not scale, is the dominant lever in our panel. Second, the bottleneck is the feature, not the classifier: at the best Qwen-7B cell a logistic gate (0.596) matches or beats an MLP-256 (0.537) and an RBF-SVM (0.535), and larger discriminators overfit the small per-setting data. This explains why a simple linear gate suffices and why the diagnostic---an AUC computed from that same linear readout---is a faithful summary of what selection can exploit.

\subsection{Matching a verifier at a fraction of the cost; cross-domain generality}\label{sec:cost}
Where it works, hidden-state fusion is inference-cheap. On OpenBioLLM/MedQA medium questions, a training-free generative self-verifier reaches 63.0\% ($+11.1$ over voting), but it requires on the order of $|\text{options}|$ extra forward passes per question (four on MedQA, cached across draws). CASE reaches 62.0\% ($+10.2$) by reusing activations already computed during generation plus a single matrix--vector product, replacing those verification passes with one matvec per candidate (Table~\ref{tab:verifier}). This is a match only in the single setting tested: the difference is $1.0$ pp on $n=48$ medium questions, without a paired equivalence test, so it establishes ``competitive at far lower inference cost,'' not general parity. The cost comparison is also inference-only. Unlike the training-free verifier, CASE carries an \emph{offline} cost that Table~\ref{tab:verifier} does not price---a labeled calibration set of a few hundred questions per (model, domain), a candidate pool, a late-layer sweep, and a gate fit---so CASE trades this one-time supervision cost for the online saving. Internal fusion is therefore best used as a low-cost complement to trained verifiers: where decodability is high it recovers most of the verifier's benefit at negligible \emph{inference} cost, and where it is low the diagnostic flags when a more expensive judge is needed.

\begin{table}[t]\centering\footnotesize
\caption{Generative verifier vs.\ CASE (OpenBioLLM $\times$ MedQA, medium difficulty). All four estimates come from the same diagnostic candidate pool and the same draws (best-of-$N$, $N=4$, 200 draws), so the majority-vote baseline (51.8\%) is shared. Gains over voting are computed from unrounded accuracies and can differ from the displayed percentages by $\pm0.1$ pp (the verifier's $+11.1$ vs.\ $63.0-51.8$). The operating point differs from the medium cell of Table~\ref{tab:main} (full-pool selection), so the CASE accuracy there, 63.2\%, differs slightly within sampling uncertainty. To quantify the cost gap: the verifier's $|\text{options}|$ extra forward passes cost on the order of $|\text{options}|\cdot 2 N_{\text{params}} T \approx 2\times10^{13}$ FLOP per question (8B model, $T\approx300$ prompt tokens), whereas CASE adds one $d{=}4096$ matrix--vector product per candidate on activations already produced during generation, $\approx10^{5}$ FLOP per question---about eight orders of magnitude less marginal compute, though both share the upstream candidate-generation cost. This is an estimate from parameter and token counts, not a wall-clock benchmark.}
\label{tab:verifier}
\begin{tabular}{llll}
\toprule
Method & accuracy & vs.\ vote & extra cost / question \\ \midrule
single agent & 46.5\% & --- & --- \\
majority vote & 51.8\% & --- & 0 \\
generative verifier & 63.0\% & $+11.1$ & $\sim|\text{options}|$ fwd passes (cached; 4 on MedQA) \\
CASE & 62.0\% & $+10.2$ & $\approx0$ (reuse activations $+$ 1 matvec) \\
\bottomrule
\end{tabular}
\end{table}

The relationship extends beyond medicine to mathematics, scored on the law's own axis (medium-bin within-question AUC and per-bin gains at the operating layer). On MATH-500 both models sit above threshold---medium-bin AUC 0.79 for the specialist Qwen2.5-Math-7B, 0.76 for the general Qwen2.5-7B---and both gain: $+8.1$ pp at medium difficulty ($p=0.043$) and $+6.6$ pp (not significant, small medium bin) respectively, with a further $+2.5$ to $+2.6$ pp on hard questions ($p<0.001$; Table~\ref{tab:math}). Mathematics thus obeys the same law. Grade-school arithmetic (GSM8K) is more ceiling-limited: both models solve $\sim$92\% of items, leaving too few non-easy questions (7 and 4 at medium difficulty) for a medium-bin estimate, so we report it on the hard subset. There the specialist is decodable (pooled AUC 0.82) and gains ($+11.9$ pp, $p=0.001$), while the general model, whose pooled decodability (0.595) sits at the 0.60 mark, does not ($+0.3$ pp, n.s.)---the gain again tracks decodability, and the specialist-over-general ordering holds. GSM8K's ceiling makes it directional support only; MATH-500, on the law's axis, is the quantitative test, and it conforms.

\begin{table}[t]\centering\footnotesize
\caption{Mathematics domains, at the operating layer as for the medical settings; $p$ from the same question-level bootstrap. For MATH-500 the AUC is the medium-bin within-question AUC (the law's axis, Eq.~\ref{eq:fit}), with medium/hard $n=17/233$ (Qwen2.5-7B) and $27/236$ (Qwen2.5-Math-7B). GSM8K is too ceiling-limited for the medium-bin axis (only 7 and 4 medium questions), so it is reported off-axis: pooled within-question AUC ($\dagger$) and hard-bin gain over $n=15$ and $18$. n.s.\ $=$ not significant.}
\label{tab:math}
\begin{tabular}{llrll}
\toprule
Model & Benchmark & within-Q AUC & medium CASE$-$vote ($p$) & hard CASE$-$vote ($p$) \\ \midrule
Qwen2.5-Math-7B & MATH-500 & 0.789 & $+8.1$ (.043) & $+2.5$ ($<$.001) \\
Qwen2.5-7B & MATH-500 & 0.763 & $+6.6$ (n.s.) & $+2.6$ ($<$.001) \\
Qwen2.5-Math-7B & GSM8K & $0.820^{\dagger}$ & --- & $+11.9$ (.001) \\
Qwen2.5-7B & GSM8K & $0.595^{\dagger}$ & --- & $+0.3$ (n.s.) \\
\bottomrule
\end{tabular}
\end{table}

Finally, we test the law outside both medicine and mathematics on graduate-level science (GPQA), a hard, knowledge-intensive benchmark on which our general models answer only $\sim$21--23\% of questions correctly, so the candidate pool is dominated by the \emph{Byzantine} hard regime. Here a general model is the natural probe: Qwen2.5-7B and -14B carry substantial latent physics and chemistry knowledge but no domain fine-tuning, letting us ask whether decodable correctness---rather than a specialist label---is what the law requires. Across the four model$\times$subfield settings the within-question decodability spans the threshold (medium-bin AUC 0.55--0.68, highest on physics), and hidden-state selection significantly beats voting on the well-powered hard (\emph{Byzantine}) subset in every case ($+6.2$ to $+7.9$ pp, all $p<0.001$; Table~\ref{tab:gpqa}, Supplementary Table~S12). The medium-difficulty gains track the law out of sample: the two physics settings, whose decodability is clearly above threshold, fall within 0.7 and 3.8 pp of the gain predicted by the medical regression (the small medium bins, $n=19$--23, preclude significance there), while the two chemistry settings sit just below the threshold with medium-difficulty gains not significantly different from zero ($-0.3$ and $-7.6$ pp, $n=28$ and 30). A general model can thus possess decodable, fusion-actionable correctness on graduate physics while lacking it on clinical medicine, supporting that the operative variable is decodability itself, not domain, specialization, or scale. Taken together with the mathematics settings, the law transfers across domains in two senses worth distinguishing. Its \emph{threshold} is domain-general: all six non-medical settings (four GPQA, two MATH-500) are classified correctly by the $\mathrm{AUC}\approx0.60$ boundary---above it they gain, below it they do not---and the GPQA gains match the medical point forecast to within RMSE $\approx3$ pp. Its \emph{slope} is not: the mathematics settings sit well above threshold and do gain, but by less than the medical regression predicts (the two MATH points fall $\sim$12 pp below the medical line), so the gain \emph{per unit} decodability is calibrated per domain. The deployable question the diagnostic answers---will internal fusion help at all?---therefore generalizes out of sample, while the exact magnitude does not, and should be recalibrated on a small labeled set in each new domain.

\begin{table}[t]\centering\footnotesize
\caption{A non-medical knowledge domain (GPQA, graduate science; general models, late-layer operating point). Decodability spans the threshold (physics above, chemistry just below), and hidden-state selection beats voting on the hard (\emph{Byzantine}) subset in every setting. The hard subset is well-powered (hard-bin $n=133$--$136$; Qwen-14B physics on its complete 187-question set), whereas the medium bins are small ($n=19$--$30$), so the medium gains are mostly not significant (n.s.\ $=$ not significant).}
\label{tab:gpqa}
\begin{tabular}{llr ccc ccc}
\toprule
& & & \multicolumn{3}{c}{medium} & \multicolumn{3}{c}{hard} \\
\cmidrule(lr){4-6}\cmidrule(lr){7-9}
Model & Subfield & AUC & vote\% & CASE\% & $\Delta$vote & vote\% & CASE\% & $\Delta$vote ($p$) \\ \midrule
Qwen2.5-14B & Phys & 0.681 & 50.3 & 61.1 & $+10.8$ (n.s.) & 3.8 & 11.6 & $+7.9$ ($<$.001) \\
Qwen2.5-7B & Phys & 0.661 & 54.0 & 58.2 & $+4.3$ (n.s.) & 4.6 & 12.0 & $+7.4$ ($<$.001) \\
Qwen2.5-14B & Chem & 0.576 & 49.9 & 49.6 & $-0.3$ (n.s.) & 6.3 & 12.6 & $+6.2$ ($<$.001) \\
Qwen2.5-7B & Chem & 0.550 & 54.9 & 47.4 & $-7.6$ (n.s.) & 6.5 & 14.2 & $+7.7$ ($<$.001) \\
\bottomrule
\end{tabular}
\end{table}

We probe what the gate encodes on GPQA, where the models sit near chance, to confirm it is the same correctness readout as in the medical case rather than a format or self-consistency artifact. Using a single general model (Qwen2.5-7B) across physics, chemistry and medicine, three properties align it with the aligned-knowledge account. First, near-chance accuracy does not preclude decodable knowledge: within-question AUC is evaluated only where correct and incorrect candidates coexist, so low accuracy merely places the correct answer in the minority; consistently, at a single common late layer used for this matched cross-domain probe, the same model is decodable where it holds knowledge (physics, within-question AUC 0.63) but not where it lacks it (clinical medicine, 0.49 $\approx$ chance); these probe-layer values differ from the operating-point AUCs in Table~\ref{tab:gpqa}. Second, the correctness direction is largely domain-general: the mass-mean correct-minus-incorrect directions across the three domains have pairwise cosine 0.52--0.83, and a gate trained on one domain ranks another's candidates above chance (cross-domain AUC 0.55--0.69; Supplementary Table~S15), so it reads a shared correct/incorrect axis rather than a subject-specific format cue. Third, the signal is not self-consistency: the gate score is slightly negatively correlated with a candidate's agreement with the model's other samples (within-question correlation $-0.07$ to $-0.10$), matching its ability to beat majority voting in the \emph{Byzantine} regime. GPQA is therefore the same phenomenon as the medical case---decodable correctness wherever the model has aligned knowledge to recall---rather than a separate mechanism.

\subsection{CASE outperforms near-free output-space selectors}\label{sec:cheap}
The appeal of internal selection is that it is almost free, which invites comparison against the equally cheap selectors that read the output distribution, rather than against an expensive verifier: self-certainty \cite{r42}, sequence log-probability (equivalently perplexity), and predictive entropy. We evaluate all of them on the same candidate pools and the same draws as CASE, computing each score from the same forward pass (Section~\ref{sec:method}). Two of the classical output-space signals collapse into baselines we already report: on multiple-choice, semantic-entropy selection groups candidates by answer option, so choosing the largest cluster is exactly majority voting; and P(True) is the generative verifier of Section~\ref{sec:cost}. Table~\ref{tab:cheap} therefore compares CASE against the three genuinely distinct near-free selectors across five settings that span the decodability threshold.

\begin{table}[t]\centering\footnotesize
\caption{CASE vs.\ near-free output-space selectors at medium difficulty, on a shared candidate pool. The self-cert / seq-logp / pred-ent columns give each selector's gain over majority voting (pp); CASE advantage is CASE minus the best of the three---a paired comparison, positive whenever CASE beats every near-free selector. within-Q AUC is the pooled within-question ROC-AUC of the CASE gate (over all difficulties on this single seed-7 pool, so it differs slightly from the operating-layer AUC in Tables~\ref{tab:main} and \ref{tab:law}). n.s.\ $=$ not significant at $\alpha=0.05$. Selector gains are over majority voting on a shared seed-7 candidate pool (operating layers L22--L27).}
\label{tab:cheap}
\begin{tabular}{lrllll}
\toprule
Model / benchmark & within-Q AUC & self-cert & seq-logp & pred-ent & \shortstack[l]{CASE\\advantage} \\ \midrule
OpenBioLLM-8B MedMCQA & 0.770 & $+4.9$ n.s. & $+13.3$ & $+13.3$ & $+3.7$ \\
OpenBioLLM-8B MedQA & 0.759 & $+0.9$ n.s. & $+14.6$ & $+13.4$ & $+9.7$ \\
OpenBioLLM-8B PubMedQA-cb & 0.707 & $-22.1$ & $-4.8$ n.s. & $-0.1$ n.s. & $+23.2$ \\
Qwen2.5-7B MedQA & 0.574 & $-3.3$ n.s. & $-3.8$ n.s. & $-3.2$ n.s. & $+4.6$ \\
meditron-7B MedQA & 0.531 & $-14.9$ & $-14.3$ & $-14.4$ & $+7.3$ \\
\bottomrule
\end{tabular}
\end{table}

Two findings stand out. First, CASE beats every near-free output-space selector in all five settings: against self-certainty the margin is 5 to 45 points at medium difficulty, and CASE leads even the strongest of the three cheap selectors in every case (Table~\ref{tab:cheap}; the full medium-and-hard breakdown is Supplementary Table~S13). The gap is largest exactly where our account predicts the output distribution to be least trustworthy. On closed-book PubMedQA---the pure parametric-recall setting---self-certainty is actively misleading, not just uninformative: its within-question AUC is 0.36 and it selects 22.1 points below voting, because a model that sounds most confident is often the one that has confabulated, whereas the internal readout still separates right from wrong. Second, below the threshold no selector helps at medium difficulty: on Qwen-7B and meditron every method, CASE included, is non-significant or negative against voting there, and every within-question AUC sits near 0.5 (the hard-bin gains in Table~\ref{tab:main} reflect the \emph{Byzantine}-failure floor of Section~\ref{sec:when}, not medium-difficulty decodability). The decodability diagnostic thus predicts a selector-independent property---whether correctness is recoverable at all---rather than merely CASE's own success, which is why the same threshold that governs CASE also governs the cheap selectors. A within-question length control confirms the advantage is not a response-length artifact: the CASE score is only weakly correlated with answer length ($\approx0.1$--0.3, against 1.0 for a pick-the-longest rule), and its hard-bin gain survives residualizing out the length-predictable component in all four settings with response-length variation (Supplementary Table~S14; meditron-7B generates to the token limit, leaving no length to control for). Its margin over the cheap selectors is nonetheless largest at medium difficulty, where the length confound is weakest. Among hidden-state selector designs on the shared pool, a LiLaVe-style shallow-tree latent verifier \cite{r49} and CASE's linear gate perform comparably ($+10.0$ vs.\ $+9.0$ pp), while a SWIFT-style token-pooled readout \cite{r50} is weaker ($+4.3$ pp); within this latent-verifier family the operative choice is the answer-token position rather than the classifier class (Supplementary Section~S14).

A final control isolates \emph{where} the signal lives: whether it is genuinely in the internal representation, or merely in the availability of a few hundred correctness labels. We trained a text classifier (TF-IDF with logistic regression) on the same question-grouped labels using the candidates' generated text rather than their hidden states, on a freshly generated OpenBioLLM/MedQA pool (150 questions, 2{,}400 candidates). With identical supervision it ranks correct from incorrect candidates within a question at chance---within-question AUC $0.44$ overall and $0.52$ at medium difficulty---against $0.70$ for the hidden-state gate on the same setting. The within-question correctness signal is therefore carried by the internal representation and is not recoverable from the answer text under the same labels: it is the states, not the labels, that do the work.

\section{A deployment recipe from the decodability law}\label{sec:recipe}
\begin{figure}[t]
\centering
\includegraphics[width=0.80\linewidth]{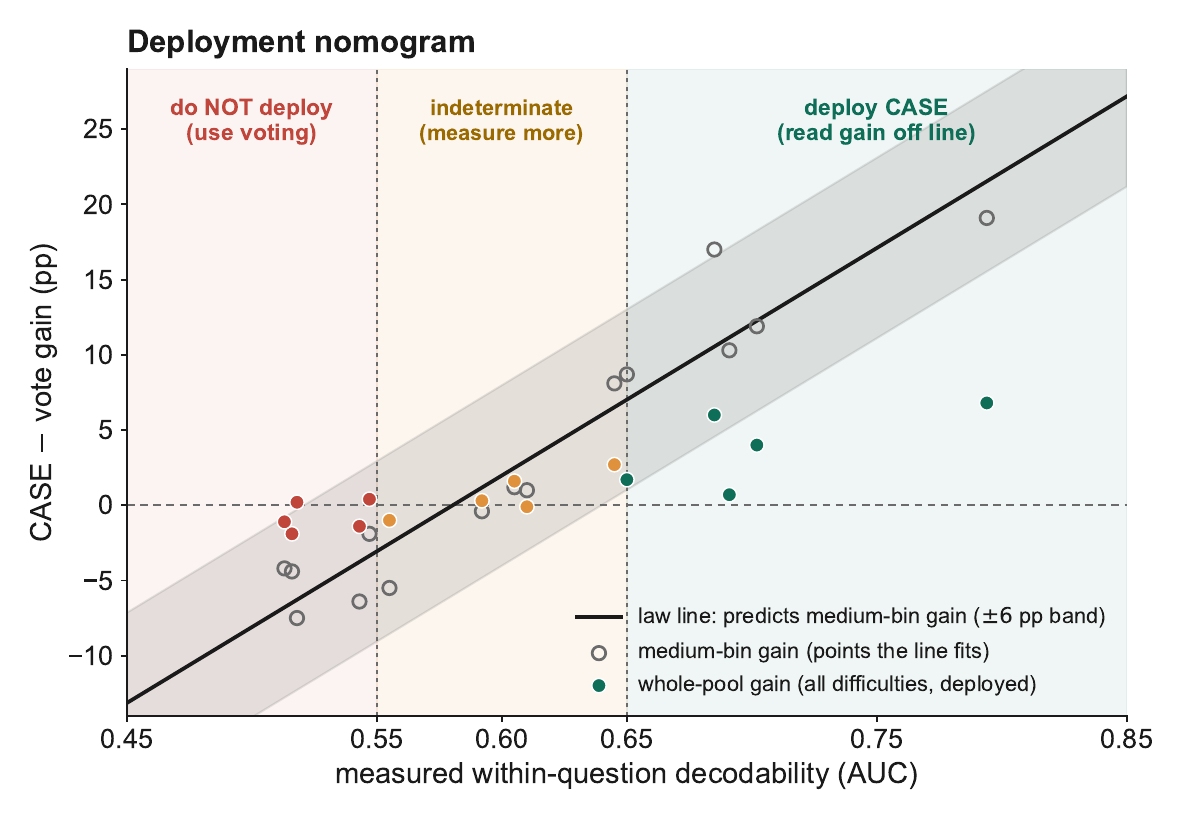}
\caption{Deployment nomogram. A practitioner measures the within-question decodability on a small calibration set, locates the AUC on the horizontal axis, and reads the expected CASE$-$vote gain and its $\pm6$ pp band off the law line. The vertical bands are the decision zones: below $\mathrm{AUC}\,0.55$ retain voting; in the indeterminate band ($0.55$--$0.65$) the measurement cannot decide; above $0.65$ deploy CASE. Open circles are the medium-difficulty gains the law is fit on (the mechanistic quantity); filled points are the difficulty-unconditional gains a deployer actually experiences (smaller, but positive exactly in the deploy zone).}
\label{fig:nomogram}
\end{figure}
The law converts a previously unpredictable heuristic into a procedure a practitioner can follow before committing to internal fusion (Fig.~\ref{fig:nomogram}). Given a target (model, domain) pair, the deployment rule is
\begin{equation}
\label{eq:deploy}
\begin{aligned}
\text{use CASE}
&\iff \AUCwq \ge a^\ast,
\qquad a^\ast \approx 0.60, \\
\text{expected gain:}\qquad
\widehat{\Delta}
&= 100.8\,\AUCwq - 58.5 .
\end{aligned}
\end{equation}
implemented by the following recipe:
\recipeitem{%
\textbf{Calibrate.} On a small labeled set (a few hundred questions suffices, and a learning curve shows the estimate is already stable at $\sim$50: held-out within-question AUC $0.72\pm0.03$ at $N_{\text{cal}}=50$ vs.\ $0.70\pm0.03$ at $400$ on OpenBioLLM/MedQA), sample a candidate pool per question, extract answer-token activations across late layers, and fit the leakage-free question-grouped gate (Section~\ref{sec:method}).}
\recipeitem{%
\textbf{Measure decodability.} Compute the within-question ranking AUC (Section~\ref{sec:diag}) at the best late layer. This single number is the diagnostic.}

\recipeitem{%
\textbf{Decide.} The AUC threshold governs medium difficulty (the rule of Eq.~\ref{eq:deploy}). Clearly above it, deploy hidden-state selection with the expected gain from the law (Eq.~\ref{eq:fit}, prediction interval $\pm6$ pp; Fig.~\ref{fig:law}); clearly below it, retain majority voting or fall back to a trained/generative verifier. Because the threshold's own confidence interval ($[0.562,0.596]$) is wider than the empirical success/failure gap, treat a band around the boundary (roughly $\mathrm{AUC}\,0.55$--$0.65$) as indeterminate and report the interval forecast in place of a binary call. Two limits keep this a guide, not a switch. First, hard, \emph{Byzantine} inputs are exempt from the threshold: where the correct answer is a minority and voting collapses, CASE beats voting even for sub-threshold models (as on graduate chemistry and meditron), exactly as Proposition~2 predicts ($a^\ast<\tfrac12$ when $V<A$), so the operator should be used on the hard regime regardless of the medium-bin AUC---though there the gain comes from avoiding the collapsed vote (any single candidate would), not from the internal signal, which improves on a single candidate only once decodability clears $\tfrac12$. Second, because difficulty is defined by the (unknown) correct-fraction, a deployer who cannot bin questions applies CASE to the whole pool and should expect the smaller \emph{unconditional} gain, which is clearly positive well above the threshold and scatters within $\pm0.4$ pp of zero near it. Fit directly on the same decodability axis, this whole-pool gain follows $\widehat{\Delta}_{\text{uncond}}=27.6\,\AUCwq-15.6$ ($r=0.873$), crossing zero at $\mathrm{AUC}=0.565$ and giving, for example, $+3.7$ pp at $\mathrm{AUC}=0.70$, so a deployer who cannot bin questions reads the expected gain off this line rather than off Eq.~\eqref{eq:deploy} (the filled points in Fig.~\ref{fig:nomogram}).}

\recipeitem{%
\textbf{Operate.} At inference, reuse the activations already produced during generation and apply the gate with one matrix--vector product per candidate; switch the operator on only in the difficulty regime where it pays (medium/hard), where the candidate pool is most likely to be \emph{Byzantine}.}

The diagnostic is cheap (it reuses the calibration pool and a linear fit), model-agnostic, and leakage-free. Its classification is robust: a leave-one-out cross-validated threshold correctly classifies 14 of 15 settings (the one error is a boundary setting near zero gain), and a single combined sign test---above/below $\mathrm{AUC}=0.60$ predicting the sign of the gain---is correct on all 15 (one-sided $p=3\times10^{-5}$, treating settings as independent; the model-cluster block bootstrap of Section~\ref{sec:law} accounts for the non-independence), so the separation does not rest on any one setting or on a table of individually underpowered comparisons (Table~\ref{tab:law}); under the stricter held-out calibration-to-deployment test it separates eleven of thirteen settings, the two exceptions lying at the threshold with near-zero gains (Section~\ref{sec:law}). It also explains failures prospectively---a general model, an open-book task, or a floor-competence specialist will measure below threshold---so a practitioner can avoid deploying internal fusion where it would silently degrade accuracy.

\section{Discussion}\label{sec:discussion}
Our findings reconcile optimistic claims that LLMs `know when they are right' with the practical fragility of hidden-state selection. Both hold, in different regimes: internal correctness is decodable and can drive a fusion rule robust to voting's \emph{Byzantine} failure---but only when the model has aligned latent domain knowledge it must recall. The within-question decodability AUC turns this condition into a single predictive number ($r=0.96$ in-sample, $0.75$ held-out) with a falsifiable threshold, replacing a heuristic with a pre-deployment test. Because CASE selects the argmax of the same gate score whose AUC is the predictor, the in-sample correlation is partly structural, and we do not rest the contribution on its magnitude. The contribution is empirical: \emph{which} (model, domain) pairs are decodable, why, and that decodability measured on one set of questions predicts the fusion gain on disjoint ones. That is an empirical map from a model's knowledge state to a deployable property, not a definitional identity---and it is the part that transfers out of sample.

Viewed through classical fusion theory, our result is a statement about where a learned combiner overcomes the accuracy--diversity limits of plain voting. Majority voting is near-optimal only when base errors are weakly correlated \cite{r37}; LLM candidates violate this because they share one model and one knowledge state, so on hard inputs their errors concentrate on the same wrong answer and voting degrades with the candidate count. A combiner that weights candidates by an internal correctness estimate can escape this regime---but only if the estimate is informative, which is exactly what the within-question decodability AUC measures. The decodability law can thus be read as a quantitative, per-(model, domain) condition for when a trainable combiner is worth using over consensus fusion, connecting a contemporary LLM phenomenon to a long-standing question in information fusion.

\textbf{Decodability, not scale or specialization, is the operative variable.} The controls make this concrete: a general model is decodable on graduate physics yet at chance on clinical medicine, while a medical specialist shows the reverse, and closed-book recall is decodable where open-book extraction from a supplied passage is not. Parameter count and a `medical' label are only proxies---meditron is a medical model that still fails, and a general model succeeds where its latent knowledge runs deep. The practical consequence is that one should not pick a model for fusion by size or domain badge but by measuring, directly, whether it holds the target knowledge decodably. Decodability is a property of the alignment between what the model knows and what the question asks, surfaced at the answer token late in the network, where the model has committed to an answer and the correctness of that commitment is most legible.

\textbf{Internal decodability is not repackaged output confidence.} The two come apart exactly in the recall regime: on closed-book PubMedQA the near-free self-certainty selector points the wrong way (Section~\ref{sec:cheap}), because the most fluent candidate is often the confabulated one, while the internal readout does not. Output-space selectors---self-consistency, predictive entropy, verbalized confidence---inherit the generator's miscalibration; an internal correctness readout can be a strictly better fusion weight precisely where parametric recall, rather than surface fluency, determines correctness. This marks the regime where cheap output-distribution signals suffice from the one where reading the residual stream is warranted.

\textbf{A readable direction is not a control lever.} Decodability does not imply steerability: across four intervention variants, adding the correctness direction to the residual stream during generation does not raise accuracy and can lower it. A direction can linearly separate correct from incorrect states---a correlate of correctness---without being a causal mediator of the answer the model emits. This cautions the representation-engineering and inference-time-intervention literature against treating any probeable direction as an actuator, and suggests the late-layer correctness signal is a downstream trace of a decision already made rather than an upstream cause of it; whether an earlier, causal correlate exists is open.

\textbf{What does this say about a model's self-knowledge?} The two dissociations, taken together, sharpen what it can mean for a language model to ``know'' whether it is right. The correctness signal is one the model can \emph{read} but not \emph{act on} (it is decodable, yet adding it back does not steer generation), and one that lives \emph{internally} even where the model's own \emph{output} points the other way (on closed-book recall, fluency and correctness anti-correlate, so the most confident-sounding answer is often the confabulated one). A model can therefore represent that an answer is likely wrong without saying so and without being able to fix it. Self-knowledge here is thus not a monolithic faculty a model possesses or lacks, but a specific, locatable property of its representations: it appears exactly where the model must recall knowledge it genuinely holds, is absent where the knowledge is shallow or supplied externally, and is legible to a simple external read-out even when it is invisible in the model's words. That gap---between what a model's states encode and what its outputs reveal---is the opening this paper exploits, and measuring it is what turns an intuition about machine self-awareness into an engineering quantity.

\textbf{Limitations.} The strongest closed-book effects are in the medical domain, where instruction-aligned models are abundant---though our GPQA results (Section~\ref{sec:cost}) show that the phenomenon and the law extend to a non-medical knowledge domain (graduate science) and to general models. Because the medical benchmarks are public, decodable correctness on them cannot be fully separated from memorized retrieval; the GPQA (deliberately search-resistant) and mathematics results, on newer or less contaminable data, are the cleaner evidence that the signal reflects usable latent knowledge rather than memorized answer keys, and the closed-book PubMedQA construction should be read with this caveat. Some settings have modest medium-difficulty sample sizes, which widen individual confidence intervals, even though the across-model law is tight (15 points, $r=0.96$; still $r=0.90$ with the entire OpenBioLLM family removed, Section~\ref{sec:law}) and predicts the out-of-sample GPQA physics gains within 3.8 pp. Our strong baseline is a generative rather than a trained process-reward verifier. The headline gains are nevertheless robust to sampling randomness: re-generating the two strongest settings with three independent seeds replicates them (medium-difficulty CASE$-$vote $=+18.1\pm1.0$ pp on MedMCQA and $+16.2\pm5.3$ pp on MedQA, significant in all six runs; Supplementary Table~S4). Given the law's tightness and its successful out-of-sample test on GPQA, we expect the AUC diagnostic to remain predictive.

\section{Conclusion}\label{sec:conclusion}
We asked when hidden-state selection should replace majority voting, and whether the answer can be known before deployment. We introduced decodability: the leakage-free within-question AUC of a linear gate that ranks a question's correct candidates above its incorrect ones. On held-out questions it predicts the gain of selection over voting at $r=0.75$, with a decision boundary near $\mathrm{AUC}=0.60$. Above that boundary CASE beats voting, and its margin grows with the number of samples on hard questions where voting collapses. Below it, voting is the better rule. Decodability tracks the aligned knowledge a model must recall, not its parameter count, and the same threshold correctly classifies mathematics and graduate-science settings that the fit never saw.

\textbf{Future directions.} Several directions follow. First, our baseline verifier is generative rather than a trained process-reward model; whether a learned verifier widens or closes the cost--accuracy gap, and whether decodability predicts \emph{its} gain too, is untested. Second, we read correctness at the final answer token on multiple-choice tasks; extending the readout to intermediate reasoning steps and to free-form generation, where `the answer token' is not sharply defined, is the main step toward general test-time fusion. Third, the stylized model predicts the threshold is not universal but set by the difficulty distribution (the sign of $V-A$), which invites testing across option counts, answer formats, and task families to map how the operating point moves. Fourth, our evidence that aligned latent knowledge \emph{causes} decodability is observational (the closed/open-book control, the physics-versus-medicine dissociation, and the necessity of the correctness direction under projection); a direct causal test, injecting domain knowledge into a general model and verifying that decodability emerges with it, is a natural next step, but requires an intervention that demonstrably raises the model's competence rather than merely shifting its output format. Finally, if correctness is a legible property of a model's states, it may be one a model can be trained to make \emph{more} legible, turning decodability from a fixed diagnostic into an optimization target and making fusion-friendliness an explicit design goal.

\section*{CRediT authorship contribution statement}

\begingroup
\setlength{\parindent}{0pt}
\setlength{\parskip}{2pt}

\textbf{Zhixiang Wang:} Conceptualization, Methodology, Software,
Investigation, Formal analysis, Writing -- original draft.

\textbf{Ziliang Hong:} Investigation, Data curation, Validation,
Visualization, Writing -- review \& editing.

\textbf{Ulas Bagci:} Conceptualization, Supervision, Funding acquisition,
Writing -- review \& editing.

\endgroup

\section*{Declaration of competing interest}
The authors declare that they have no known competing financial interests or personal relationships that could have appeared to influence the work reported in this paper.

\section*{Data availability}
All benchmarks used (LogiQA, MedQA, MedMCQA, PubMedQA, MATH-500, GSM8K, GPQA) are publicly available, and all models are open-weight and obtained from their public releases. The extracted hidden-state features, per-run logs, and derived result tables will be deposited in a public repository upon acceptance; a link is provided in the code release below.

\section*{Code availability}
An anonymized repository containing the code to generate candidates, extract answer-token activations, fit the leakage-free gate, compute the within-question decodability diagnostic, and reproduce every figure and table is available for peer review at \url{https://anonymous.4open.science/r/case-decodability-28D1/}; it will be de-anonymized and released publicly under an open-source license upon acceptance.

\section*{Acknowledgments}
This study is partially supported by NIH grants R01-HL171376 and U01-CA268808.

\fontsize{7.5pt}{8.5pt}\selectfont
\sloppy

\end{document}